\documentclass[Afour,sageh,times]{sagej}
\usepackage{natbib}
\usepackage{stfloats}
\usepackage{moreverb,url}
\usepackage{graphicx}
\usepackage{xcolor}
\usepackage{soul}  % for \hl
\sethlcolor{yellow!20} 
\usepackage{subcaption}
\usepackage{multirow}
\usepackage{diagbox}
\usepackage{hhline}
\usepackage[flushleft]{threeparttable}
\usepackage[justification=centering]{caption}
\usepackage[colorlinks,bookmarksopen,bookmarksnumbered,citecolor=red,urlcolor=red]{hyperref}

\newcommand\normx[1]{\left\Vert#1\right\Vert}
\DeclareMathOperator*{\argmax}{argmax}
\DeclareMathOperator*{\argmin}{argmin}

\newcommand\BibTeX{{\rmfamily B\kern-.05em \textsc{i\kern-.025em b}\kern-.08em
T\kern-.1667em\lower.7ex\hbox{E}\kern-.125emX}}

\def\volumeyear{2025}

\begin{document}

\runninghead{Yang et al.}

\title{Active 6D Pose Estimation for Textureless Objects using Multi-View RGB Frames}

\author{Jun Yang\affilnum{1,2}, Wenjie Xue\affilnum{2}, Sahar Ghavidel\affilnum{2} and Steven L. Waslander\affilnum{1}}

\affiliation{\affilnum{1}University of Toronto Institute for Aerospace Studies and Robotics Institute, ON, Canada\\
\affilnum{2}Epson Canada Ltd, ON, Canada\\
This work was supported by Epson Canada Ltd.}

\corrauth{Jun Yang, University of Toronto Institute for Aerospace Studies,
M3H~5T6, ON, Canada.}

\email{jun.yang@robotics.utias.utoronto.ca}

\begin{abstract}
Estimating the 6D pose of textureless objects from RGB images is an important problem in robotics. Due to appearance ambiguities, rotational symmetries, and severe occlusions, single-view based 6D pose estimators are still unable to handle a wide range of objects, motivating research towards multi-view pose estimation and next-best-view prediction that addresses these limitations. In this work, we propose a comprehensive active perception framework for estimating the 6D poses of textureless objects using only RGB images. Our approach is built upon a key idea: decoupling the 6D pose estimation into a two-step sequential process can greatly improve both accuracy and efficiency. First, we estimate the 3D translation of each object, resolving scale and depth ambiguities inherent to RGB images. These estimates are then used to simplify the subsequent task of determining the 3D orientation, which we achieve through canonical scale template matching. Building on this formulation, we then introduce an active perception strategy that predicts the next best camera viewpoint to capture an RGB image, effectively reducing object pose uncertainty and enhancing pose accuracy. We evaluate our method on the public ROBI and TOD datasets, as well as on our reconstructed transparent object dataset, T-ROBI. Under the same camera viewpoints, our multi-view pose estimation significantly outperforms state-of-the-art approaches. Furthermore, by leveraging our next-best-view strategy, our approach achieves high pose accuracy with fewer viewpoints than heuristic-based policies across all evaluated datasets. The accompanying video and T-ROBI dataset will be released on our project page: \url{https://trailab.github.io/ActiveODPE}.
\end{abstract}


\maketitle

\section{Introduction}
Texture-less rigid objects occur frequently in industrial environments and are of significant interest in many robotic applications. The task of 6D pose estimation aims to detect objects of known geometry and estimate their 6DoF (Degree of Freedom) poses, i.e., 3D translations and 3D orientations, with respect to a global coordinate frame. In robotic manipulation tasks, accurate object poses are required for path planning and grasp execution~\citep{song2017cad, tremblay2018deep, wang2019densefusion, deng2020self}. For robotic navigation, 6D poses serve as valuable cues for localization and obstacle avoidance~\citep{salas2013slam++, fu2021multi, merrill2022symmetry, wang2021dsp, liao2024uncertainty}.

Due to the absence of appearance features, 6D pose estimation for textureless objects has typically been addressed using depth data~\citep{drost2010model, bui2018regression, gao20206d, gao2021cloudaae, yang2021probabilistic, cai2022ove6d, li2023depth} or RGB-D images~\citep{doumanoglou2016recovering, wang2019densefusion, wada2020morefusion, he2020pvn3d, tian2020robust, wen2020se, saadi2021optimizing, li2023multi, wen2024foundationpose}. These methods demonstrate strong pose estimation performance when high-quality depth data is available. However, despite advances in depth sensing technology, commodity-grade depth cameras frequently produce inaccurate depth maps, with errors or missing data occurring on glossy or dark surfaces~\citep{chai2020deep, yang2022next, yang2024active}, as well as on translucent or transparent objects~\citep{sajjan2020clear, liu2020keypose, xu2021seeing}. These depth limitations can severely degrade object pose estimation performance. Therefore, RGB-based approaches have received a lot of attention over the past decade as a promising alternative~\citep{hinterstoisser2011gradient, brachmann2016uncertainty}. 

Due to the numerous advances in deep learning over the last decade, some learning-based approaches have recently been shown to significantly improve object pose estimation performance using only RGB images~\citep{kehl2017ssd, xiang2018posecnn, sundermeyer2018implicit, li2018deepim, peng2019pvnet, hodan2020epos, labbe2022megapose, he2023contourpose, xu20246d, sun2025metal}. However, due to the inherent scale, depth, and perspective ambiguities from a single viewpoint, RGB-based solutions often suffer from low accuracy in the final 6D pose estimation. To this end, recent works leverage multiple RGB views to enhance their pose estimation results~\citep{labbe2020cosypose, deng2021poserbpf, shugurov2021multi, fu2021multi, maninis2022vid2cad, merrill2022symmetry, labbe2022megapose, haugaard2023multi}. Although fusing multi-view information can enhance overall performance, addressing challenges such as appearance ambiguities, rotational symmetries, and occlusions remains difficult. Additionally, even when multi-view fusion mitigates some of these issues, relying on a large number of viewpoints is often impractical for many real-world applications, such as robotic manipulation.

\begin{figure}[t]
\centering
\begin{subfigure}{0.47\textwidth}
  \centering
  \includegraphics[width=\linewidth]{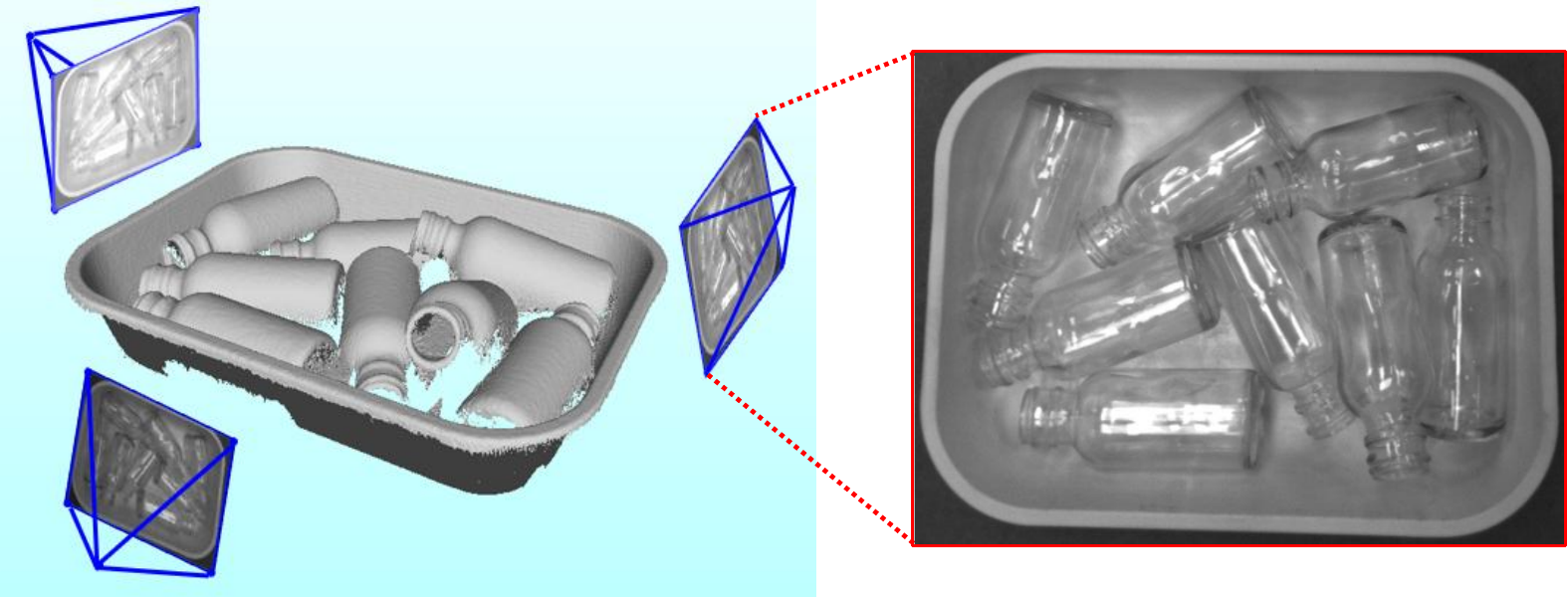}  
  \caption{Input: Multi-View RGB Images.}  
\end{subfigure}
\begin{subfigure}{0.235\textwidth}
    \centering
    \includegraphics[width=\linewidth]{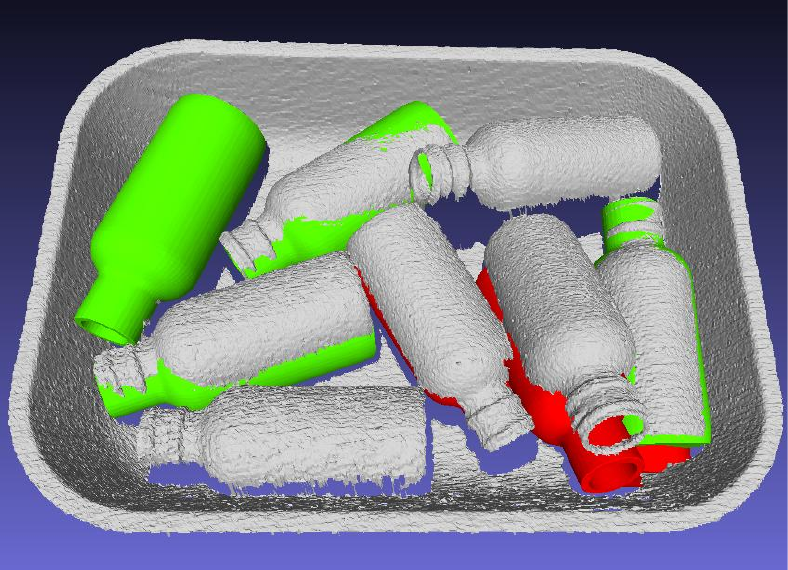}    
    \caption{Results from CosyPose.}      
\end{subfigure}
\begin{subfigure}{0.235\textwidth}
    \centering
    \includegraphics[width=\linewidth]{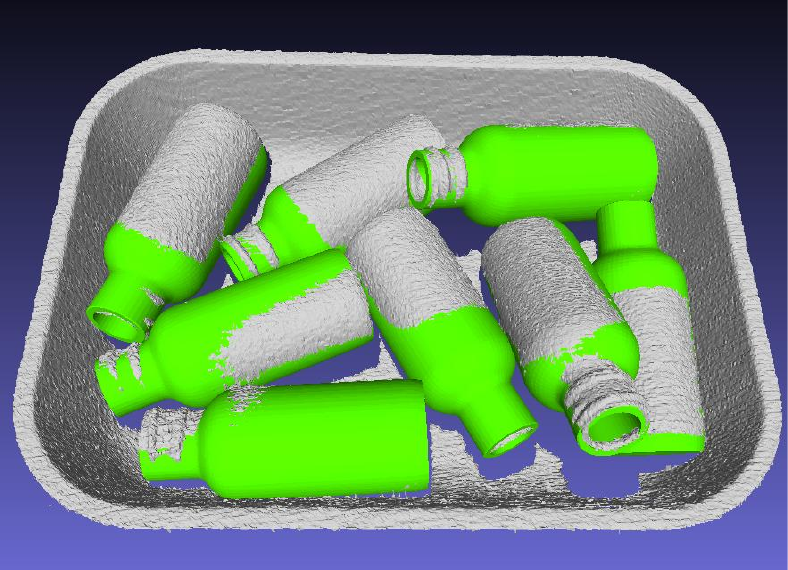}    
    \centering
    \caption{Results from Ours.}      
\end{subfigure}
\caption{6D object pose estimation using multi-view acquired RGB images. (a) The input multi-view RGB images with known camera poses. (b) The pose estimation results using CosyPose and PVNet. (c) The pose estimation results using our approach. The green and red colors represent correct and incorrect pose estimations, respectively.}
\label{fig_vis}
\end{figure}

To address these challenges, we present a comprehensive framework for both object pose estimation and next-best-view prediction using multi-view RGB images. First, we introduce a multi-view object pose estimation method that decouples the 6D pose estimation into a two-step sequential process: we first estimate the 3D translation, followed by the 3D orientation of each object. This decoupled formulation first resolves scale and depth ambiguities from single RGB images, and then leverages the resulting translation estimates to simplify object orientation estimation in the second stage. To address the multimodal nature of orientation space, we develop an optimization scheme that accounts for object symmetries and counteracts measurement uncertainties. The second part of our framework focuses on next-best-view (NBV) prediction, which builds upon the proposed multi-view pose estimator. We introduce an information-theoretic approach to quantify object pose uncertainty. In each NBV iteration, we predict the expected object pose uncertainty for each potential viewpoint and select the next camera viewpoint that minimizes this uncertainty, ensuring more informative RGB measurements are collected. Figure~\ref{fig_vis} illustrates the effectiveness of our multi-view approach on transparent objects, demonstrating accurate 6D pose estimations even under challenging conditions.

We conduct extensive experiments on the public ROBI~\citep{yang2021robi} and TOD~\citep{liu2020keypose} datasets, as well as on a challenging transparent object dataset, T-ROBI, that we present. To support network training, we also propose a large-scale synthetic dataset based on both ROBI and T-ROBI. Our approach significantly outperforms state-of-the-art RGB-based methods. Compared to depth-based methods, it achieves comparable performance on reflective objects and fully surpasses them on transparent objects, despite relying solely on RGB images. Furthermore, compared to baseline viewpoint selection strategies, our next-best-view strategy achieves high object pose accuracy while requiring fewer viewpoints.

Our work makes the following key contributions.
\begin{itemize}
    \item We propose a novel 6D object pose estimation framework that decouples the problem into a two-step sequential process. This process resolves the depth ambiguities from RGB frames and greatly improves the estimate of orientation parameters.
    \item Building on our proposed pose estimator, we introduce an information-theoretic active vision strategy that optimizes object pose accuracy by selecting the next-best camera viewpoint. 
    \item We introduce a multi-view dataset of transparent objects, specifically designed to evaluate 6D pose estimation for transparent parts in cluttered and occluded bin scenarios.
    \item To support network training, we create a large-scale synthetic dataset that includes all parts from both the public ROBI and our T-ROBI dataset.
\end{itemize}

It is important to note that this work substantially extends our previous conference paper~\citep{yang20236d}, as follows:
\begin{itemize}
    \item \textbf{Improved Orientation Estimation.} To enhance object orientation estimation, we introduce a new head into the neural network architecture that extracts per-frame object edge maps, serving as more accurate and consistent shape inputs for the object orientation estimator.
    \item \textbf{Active Vision.} We extend our previous approach by integrating an active vision strategy that selects the next-best-view to improve the object pose accuracy. 
    \item \textbf{Transparent Object Dataset.} The dataset allows evaluation of our method under real-world, challenging scenarios, while also serving as a valuable benchmark for researchers working on transparent object pose estimation.
    \item \textbf{Synthetic Dataset.} We generate a large-scale synthetic dataset to provide a comprehensive benchmark for training and fair comparison on ROBI and our transparent object dataset.
    \item \textbf{Expanded Real-World Results.} We include detailed ablation analysis, three additional baselines, and more extensive real-world results on the ROBI, TOD and our transparent object datasets.
\end{itemize}

The rest of the paper is organized as follows. Section~\ref{sec2} reviews the relevant literature. Section~\ref{sec3} formulates the multi-view object pose estimation problem. Section~\ref{sec4} describes the next-best-view prediction approach. Section~\ref{sec5} presents the experimental results, and Section~\ref{sec6} concludes the paper.

\section{Related Works}
\label{sec2}
\subsection{Object Pose Estimation from a Single RGB Image}
\label{sec2a}
\textbf{Traditional Methods.} Due to the lack of appearance features, traditional methods usually tackle the problem via holistic template matching techniques~\citep{hinterstoisser2011gradient,imperoli2015d}, but are susceptible to failure due to scale change and occlusion. Later advances~\citep{brachmann2016uncertainty} improved the efficiency of template matching by jointly regressing object coordinates and labels via a learning-based framework, but its accuracy is far below modern deep learning methods.

\textbf{End-to-End Methods.} With advances in deep learning, many works leverage convolution neural networks (CNNs)~\citep{kehl2017ssd, xiang2018posecnn, li2018deepim, wang2021gdr, labbe2022megapose} or vision transformers (ViTs)~\citep{amini2021t6d, jantos2023poet} to estimate object pose end-to-end. Among these, SSD-6D~\citep{kehl2017ssd} was the first to regress 6D object pose from a single RGB image with CNNs. To avoid complex rotation parametrization, it discretizes rotation as a classification problem. PoseCNN~\citep{xiang2018posecnn} improved this by decoupling translation and rotation with separate branches, leading to more accurate estimates. Building on these, DeepIM~\citep{li2018deepim} iteratively refines object poses by matching rendered object image to the input image. To better facilitate end-to-end methods, a continuous 6D rotation representation~\citep{zhou2019continuity} was introduced, offering advantages over other parametrizations for network training. This representation is subsequently adopted in end-to-end works~\citep{wang2021gdr, amini2021t6d, labbe2022megapose, jantos2023poet}, further improving 6D pose estimation.

\textbf{Learning-based Indirect Methods.} While classical feature and geometric-fitting methods fail on texture-less objects, deep learning overcomes this by learning discriminative features. Recent indirect methods use deep networks to predict 2D object keypoints~\citep{rad2017bb8, pavlakos20176, peng2019pvnet, he2023contourpose} or dense 2D–3D correspondences~\citep{zakharov2019dpod, park2019pix2pose, hodan2020epos, haugaard2022surfemb, su2022zebrapose}, then compute poses via RANSAC/PnP~\citep{lepetit2009epnp}. More recently, the representation power of diffusion~\citep{ho2020denoising} and foundation models~\citep{cherti2023reproducible, oquab2023dinov2} has further improved indirect methods~\citep{xu20246d}, enabling zero-shot 6D pose estimation~\citep{ausserlechner2024zs6d, fan2024pope, deng2025pos3r, sun2025metal}.

Although these methods perform well on 2D metrics, their 6D pose accuracy is limited by depth ambiguities and occlusions from a single viewpoint. Consequently, depth data is often needed to refine object poses~\citep{deng2021poserbpf, zhang2019fast, yang2024active}.

\subsection{Object Pose Estimation from Multi-View RGB Images}
Multi-view approaches address the scale and depth ambiguities that commonly occur in single-viewpoint scenarios, improving the accuracy of estimated poses. Traditional methods rely on local features~\citep{eidenberger2010active, collet2010efficient} but struggle to handle textureless objects. More recently, multi-view object pose estimation has been revisited with neural networks. These approaches employ an offline, batch-based optimization framework, where all frames are processed simultaneously to produce a consistent interpretation of the scene~\citep{kundu2018object, labbe2020cosypose, liu2020keypose, shugurov2021multi,haugaard2023multi, chen2024multi}. The most notable work is CosyPose~\citep{labbe2020cosypose}, which integrates single-view pose estimates into a globally consistent scene and is agnostic to the choice of pose estimator. Using a similar multi-view setup, a pose refiner further improves accuracy via differentiable rendering~\citep{shugurov2021multi}.

Other approaches address multi-view pose estimation in an online manner. 6D object pose tracking~\citep{deng2020self, deng2021poserbpf, labbe2022megapose, moon2024genflow} focuses on estimating object poses relative to the camera, whereas object-level SLAM simultaneously estimates both camera and object poses within a shared world coordinate frame~\citep{yang2019cubeslam, fu2021multi, wu2020eao, merrill2022symmetry, chen2024multi}. PoseRBPF~\citep{deng2021poserbpf} represents an early effort in online 6D object pose tracking, combining particle filtering with deep neural networks to achieve robust estimation under challenging conditions. In contrast, MegaPose~\citep{labbe2022megapose} and GenFlow~\citep{moon2024genflow} uses an end-to-end framework, enabling 6D tracking of novel, previously unseen objects. While these methods handle single objects well, they cannot track multiple objects simultaneously. Object-level SLAM approaches~\citep{yang2019cubeslam, fu2021multi, wu2020eao, merrill2022symmetry}, on the other hand, can recover the poses of multiple objects at once, offering a more comprehensive understanding of the scene.

While the above methods improve performance using only RGB images, they still face challenges in handling object scales, rotational symmetries, and measurement uncertainties. Our approach follows the principles of online object-level SLAM, but with the known camera poses. Using per-frame neural network predictions as measurements, our approach resolves depth and scale ambiguities through a two-step sequential formulation. It also handles rotational symmetries and measurement uncertainties within an incremental online framework.
\\

\subsection{Pose Uncertainty and Visual Ambiguity}
\label{ambiguity}
In practice, the estimated object pose may carry state uncertainty or be subject to visual ambiguity. Representing these factors is crucial for many robotic applications, such as manipulation~\citep{deng2020self, wang2019densefusion} or navigation~\citep{salas2013slam++, fu2021multi}. Object pose uncertainty reflects variability in translation and orientation, potentially with different variances along each axis (e.g., depth uncertainty along the optical axis from a single viewpoint). A straightforward strategy is to assume a unimodal distribution and represent the pose with a single covariance matrix. To estimate this covariance, several works~\citep{peng2019pvnet, richter2019towards, merrill2022symmetry, yang2023object} adopt a structured strategy that first computes the uncertainty of 2D keypoint detections and then propagates it to the 6D pose. Extending this idea to depth information, studies such as~\citep{salas2013slam++, he2020pvn3d, yang2024active, liao2024uncertainty} predict depth or 3D keypoint uncertainty and propagate it to the pose as well. Under the unimodal assumption, the resulting covariance can be further reduced by incorporating additional measurements.

Although the unimodal assumption is effective, it may fail to capture complex uncertainties, particularly when objects appear similar under different poses due to shape symmetries, occlusion, or repetitive textures, which is also known as visual ambiguities~\citep{manhardt2019explaining, bui20206d, okorn2020learning, deng2022deep, hofer2023hyperposepdf, hsiao2024confronting}. It is important to model such visual ambiguities using more expressive, multimodal distributions. Furthermore, when distinctive object features are visible, the distribution should naturally converge to a more confident, unimodal estimate. For this purpose, sampling-based methods~\citep{shi2021fast, kendall2016modelling, haugaard2023spyropose} generate multiple pose hypotheses to estimate this uncertainty, but achieving high accuracy requires many samples, making these approaches computationally costly. Alternatively, Manhardt et al.~\citep{manhardt2019explaining} learn orientation distributions using a Winner-Takes-All (WTA) strategy~\citep{rupprecht2017learning} over multiple hypotheses. To capture full orientation distributions, AAE~\citep{sundermeyer2018implicit} and PoseRBPF~\citep{deng2021poserbpf} adopt a discrete representation, whereas Okorn et al.~\citep{okorn2020learning} use a histogram-based approach. While effective, these methods are inherently limited by discretization. To address this limitation, some approaches employ mixture of Gaussian~\citep{fu2021multi} or Bingham mixture models~\citep{gilitschenski2019deep, deng2022deep} to represent multimodal orientation distributions.

In this work, we focus on modeling pose uncertainty rather than explicitly resolving visual ambiguities, enabling more efficient representations for robotic tasks.

\subsection{Active Vision}
Active vision \citep{aloimonos1988active,chen2011active,bajcsy2018revisiting}, or more specifically Next-Best-View (NBV) prediction~\citep{connolly1985determination}, refers to actively manipulating the camera viewpoint to obtain the maximum information in the next frame for the required task. Active vision has received a lot of attention from the robotics community and has been employed in many applications, such as robot manipulation~\citep{morrison2019multi,breyer2022closed,fu2024low}, calibration~\citep{rebello2017autonomous,yang2023next,choi2023accurate,xu2023observability}, object pose estimation~\citep{eidenberger2010active,wu2015active,doumanoglou2016recovering,sock2020active,yang2024active}, 3D reconstruction~\citep{isler2016information,forster2014appearance,yang2022next} and localization~\citep{davison2002simultaneous,zhang2018perception,falanga2018pampc,zhang2019beyond,hanlon2024active}. The next-best-view selection is often achieved by finding the viewpoint that maximizes the information gain or minimizes the expected entropy~\citep{rebello2017autonomous,yang2023next,choi2023accurate,xu2023observability,doumanoglou2016recovering,zhang2018perception,kiciroglu2020activemocap,zhang2019beyond,yang2024active}. To estimate 6D object poses, Doumanoglou et al. first present a single-shot object pose estimation approach based on Hough Forests~\citep{doumanoglou2016recovering}. The next-best-view is predicted by exploiting the capability of Hough Forests to compute the entropy. To eliminate reliance on the Hough Forests, recent studies show that the next-best-view can be achieved by maximizing the Fisher information of the robot state parameters~\citep{forster2014appearance,rebello2017autonomous,zhang2018perception,zhang2019beyond,yang2023next,yang2024active}. For example, in robot localization, the authors in~\citep{zhang2018perception,zhang2019beyond} use the Fisher information maximization to find highly informative trajectories and achieve high localization accuracy. Our approach lies in extending this principle to 6D object pose estimation. We actively move the camera to maximize the Fisher information, selecting viewpoints that most effectively reduce object pose uncertainty.

\begin{figure*}[t]
\centering
  \parbox{0.97\linewidth}{%
    \centering
    \includegraphics[width=\linewidth]{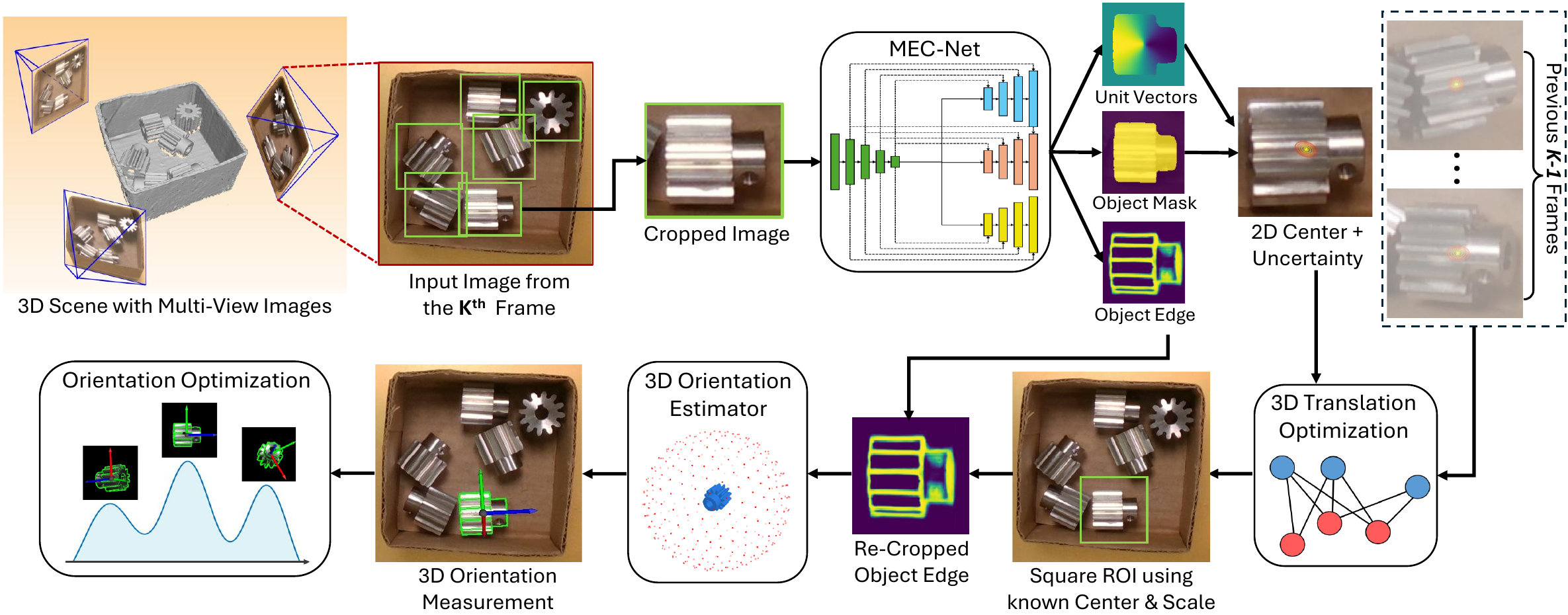}
    \caption{An overview of the proposed multi-view object pose estimation pipeline with a two-step optimization formulation. We decouple the 6D pose estimation into a sequential process: first estimating the 3D translation, then the 3D orientation of each object. At each viewpoint, SEC-Net predicts the object's 2D center, mask, and edge map. Using observations from $K$ frames, we estimate translation via multi-view optimization, which provides object scale and center for re-cropping the edge map. Per-frame orientations are then obtained and fused via a max-mixture optimization.}
    \label{fig_pipeline_rgb}
}
\end{figure*}

\section{6D Pose Estimation using Multi-View Optimization}
\label{sec3}
\subsection{Problem Formulation}
\label{sec_pose}

Given a 3D object model and multi-view images, the goal of 6D object pose estimation is to estimate the rigid transformation $\boldsymbol{T}_{wo}$ that maps the object model frame ${O}$ to a global (world) frame ${W}$, where
\begin{equation}
\boldsymbol{T}_{wo} =
\begin{bmatrix}
\mathbf{R}_{wo} & \mathbf{t}_{wo} \\
\mathbf{0}^\top & 1
\end{bmatrix} \in \mathbb{SE}(3),
\end{equation}
with $\mathbf{R}_{wo} \in \mathbb{SO}(3) \;\text{and}\; \mathbf{t}_{wo} \in \mathbb{R}^3$. For multi-view object pose estimation, we assume that the camera poses $\boldsymbol{T}_{wc} \in \mathbb{SE}(3)$ relative to the world frame are known. These can be determined through robot forward kinematics and eye-in-hand calibration~\citep{tsai1989new} when the camera is mounted on the end-effector of a robotic arm, or through off-the-shelf SLAM methods~\citep{klein2007parallel, mur2015orb} for hand-held cameras. 

Given measurements $\mathbf{Z}_{1:k}$ up to viewpoint $k$, we aim to estimate the posterior distribution of the 6D object pose $P\left(\mathbf{R}_{wo}, \mathbf{t}_{wo}|\mathbf{Z}_{1:k}\right)$. Direct computation of this distribution is typically infeasible because object translation $\mathbf{t}_{wo}$ and orientation $\mathbf{R}_{wo}$ follow distinct distributions. In particular, the translation distribution $P\left(\mathbf{t}_{wo}\right)$ is generally unimodal in practice. Although a multimodal formulation could be used to address more challenging situations, such as severe occlusions, we adopt a unimodal model to maintain computational efficiency. In contrast, the distribution for object orientation $P\left(\mathbf{R}_{wo}\right)$ is more complex, influenced by uncertainties related to object shape symmetries, appearance ambiguities, and potential occlusions. Inspired by \citep{deng2021poserbpf}, we decouple the pose posterior $P\left(\mathbf{R}_{wo}, \mathbf{t}_{wo}|\mathbf{Z}_{1:k}\right)$ into:
\begin{equation}
\label{equ_formulation_rgb}
    P\left(\mathbf{R}_{wo}, \mathbf{t}_{wo}|\mathbf{Z}_{1:k}\right) = P\left(\mathbf{R}_{wo}|\mathbf{Z}_{1:k}, \mathbf{t}_{wo}\right) P\left(\mathbf{t}_{wo}|\mathbf{Z}_{1:k}\right),
\end{equation}
where $P\left(\mathbf{t}_{wo}|\mathbf{Z}_{1:k}\right)$ can be formulated as a unimodal Gaussian distribution, $\mathcal{N}\left(\mathbf{t}_{wo}|\mathbf{\mu},\boldsymbol{\Sigma}\right)$ and  $P\left(\mathbf{R}_{wo}|\mathbf{Z}_{1:k}, \mathbf{t}_{wo}\right)$ is the orientation distribution conditioned on the input images $\mathbf{Z}_{1:k}$ and the 3D translation $\mathbf{t}_{w,o}$. To represent the complex orientation uncertainties, we follow a similar approach to \citep{eidenberger2010active} and model $P\left(\mathbf{R}_{wo}|\mathbf{Z}_{1:k}, \mathbf{t}_{wo}\right)$ as a mixture of Gaussian distributions:

\begin{equation}
\label{equ_rot_dist}
    P\left(\mathbf{R}_{wo}|\mathbf{Z}_{1:k}, \mathbf{t}_{wo}\right) = \sum_{i=1}^{N}w_i \mathcal{N}\left(\mathbf{R}_{wo}|\boldsymbol{\mu}_i,\boldsymbol{\Sigma}_i\right),
\end{equation}
which consists of $N$ Gaussian components, where $w_i$ is the weight of the $i^{th}$ component, and $\boldsymbol{\mu}_i$ and $\boldsymbol{\Sigma}_i$ are its mean and covariance, respectively.

Our decoupling formulation implies a useful correlation between the object's translation and orientation in the image domain. The 3D translation estimation $\mathbf{t}_{wo}$ is independent of the object's orientation and encodes its center and scale information. By applying the camera pose $\mathbf{T}_{wc, k}$ at frame $k$, the estimated 3D translation $\mathbf{t}_{co, k}$ under the camera coordinate provides the scale and 2D center of the object in the image. Using these estimates, the per-frame object orientation measurement $\mathbf{R}_{co, k}$ can be estimated from its visual appearance in the image. With this sequential process, our multi-view framework consists of two main steps, as summarized in Figure~\ref{fig_pipeline_rgb}. 

To implement this formulation, a key step in our framework is estimating per-frame measurements using a neural network, which are then integrated into our two-step sequential process. The network outputs the object's 2D edge map, segmentation mask, and the 2D projection of its 3D center. We refer to our network as \textbf{MEC-Net} (Mask, Edge, Center Network) for the remainder of this paper. With these estimates, we proceed to the first step (Section~\ref{Sec:translation}), where we estimate the 3D translation $\mathbf{t}_{wo}$ by minimizing the 2D re-projection error across camera viewpoints. Using the estimated 3D translation $\mathbf{t}_{wo}$, the second step (Section~\ref{Sec:rotation}) involves re-cropping an orientation-independent Region of Interest (RoI) from each object's edge map, based on the estimated scale. This RoI is then fed into an orientation estimator~\citep{hinterstoisser2011gradient} to obtain the per-frame 3D orientation measurement $\mathbf{R}_{co, k}$. The final object orientation $\mathbf{R}_{wo}$ is determined through an optimization approach that explicitly accounts for shape symmetries and incorporates a max-mixture formulation \citep{olson2013inference, fu2021multi} to mitigate uncertainties arising from per-frame orientation estimates.

\subsection{3D Translation Estimation}
\label{Sec:translation}
\begin{figure}[t]
\centering
  \parbox{0.95\linewidth}{
    \centering
    \includegraphics[width=\linewidth]{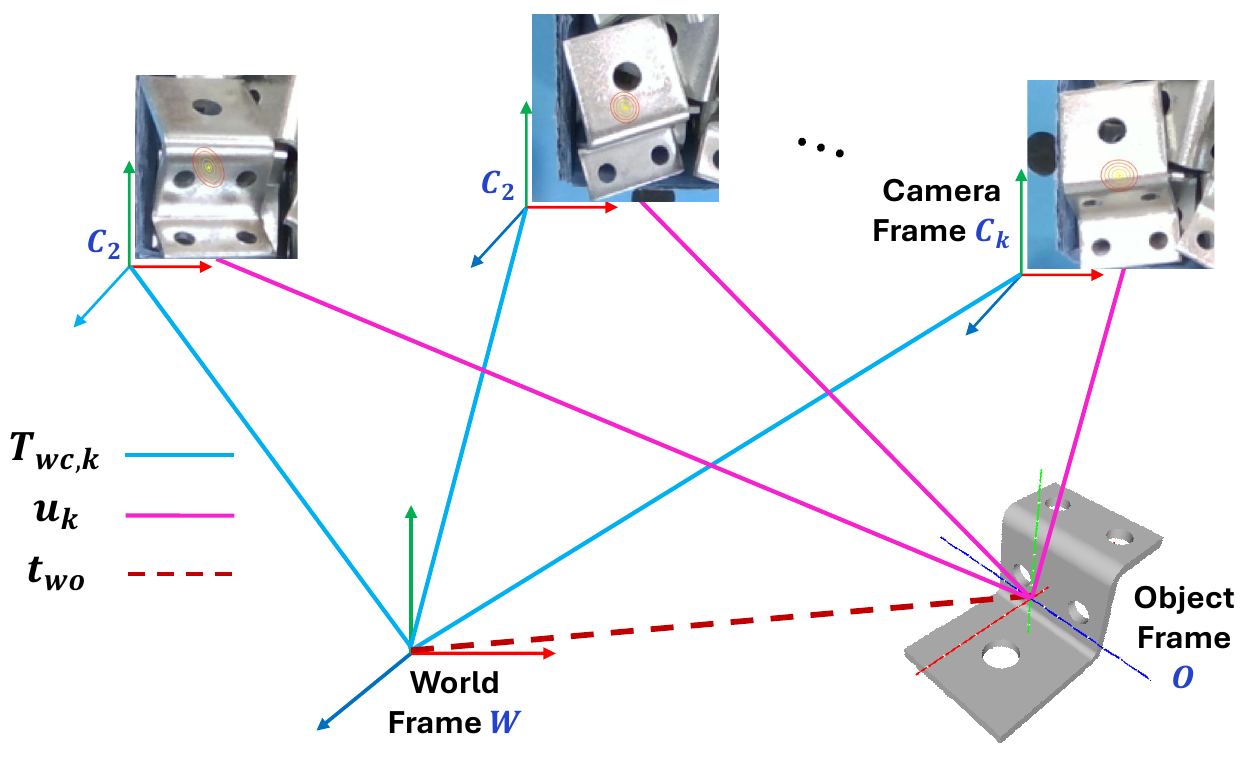}\\
    \caption{Object, world, and camera coordinate frames. 
    The 3D translation $\mathbf{t}_{wo}$ is the object origin in world coordinates, estimated by localizing per-frame 2D centers $\mathbf{u}_{k}$ and minimizing re-projection errors with known camera poses $\mathbf{T}_{wc,k}$.}
    \label{fig_omp_trans}
  }
\end{figure}

As illustrated in Figure~\ref{fig_omp_trans}, the 3D translation $\mathbf{t}_{wo}$ represents the coordinate of the object model origin in the world frame. Given that the camera pose $\mathbf{T}_{wc}$ is known, this is equivalent to solving for the translation from the object model origin to the camera optical center, $\mathbf{t}_{co} = {\left[t_x, t_y, t_z \right]}^T$. Given an RGB image from an arbitrary camera viewpoint, the translation $\mathbf{t}_{co}$ can be recovered by the following back-projection, assuming a pinhole camera model,
\begin{equation}
\label{equ3}
    \mathbf{t}_{co}
    =
    \begin{bmatrix}
        t_x \\
        t_y \\
        t_z
    \end{bmatrix}
    =
    \begin{bmatrix}
        \frac{u_x - c_x}{f_x} t_z \\
        \frac{u_y - c_y}{f_y} t_z \\
        t_z
    \end{bmatrix}, 
\end{equation}
where $f_x$ and $f_y$ denote the camera focal lengths, and ${\left[c_x, c_y\right]}^T$ is the principal point. We define $\mathbf{u}={\left[u_x, u_y\right]}^T$ as the 2D projection of the object model origin $O$ and refer to it as the 2D center of the object in the rest of the paper. If the object center $\mathbf{u}$ is localized in the image and the depth $t_z$ to the object center is estimated, then $\mathbf{t}_{co}$ (or $\mathbf{t}_{wo}$) can be recovered. In our framework, we use MEC-Net to predict the 2D object center $\mathbf{u}$ for each frame and estimate the depth $t_z$ through a multi-view optimization formulation.

Our MEC-Net architecture is shown in the upper part of Figure \ref{fig_pipeline_rgb} and is based on PoseCNN~\citep{xiang2018posecnn} and PVNet~\citep{peng2019pvnet}. To handle multiple instances within the scene, we first employ YOLOv8~\citep{sohan2024review} to detect 2D bounding boxes of the objects. These detections are then cropped and resized to 128x128 before being passed to the network. To estimate the object 2D center, the MEC-Net first predicts pixel-wise binary labels and a 2D vector field towards the object center. A RANSAC-based voting scheme is then applied to compute the mean $\mathbf{u}_k$ and covariance $\boldsymbol{\Sigma}_{\mathbf{u},k}$ of the object center at frame $k$. For more details on object center estimation, we refer the reader to~\citep{xiang2018posecnn, peng2019pvnet}.

\begin{figure*}[t]
\centering
  \includegraphics[width=\linewidth]{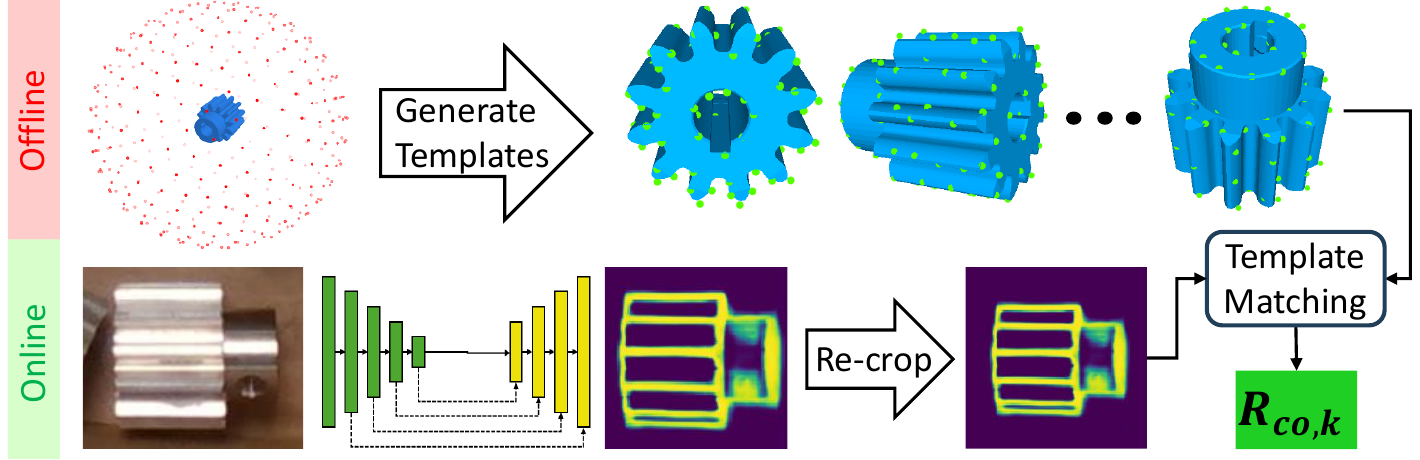}
\caption{The process of acquiring the per-frame object orientation measurement, $\mathbf{R}_{co, k}$. This process is based on a template matching (TM)-based approach, LINE-2D. To reduce the gap between rendered templates and RoI images, we introduce a new head into our neural network, MEC-Net (shown in the upper part of Figure~\ref{fig_pipeline_rgb}), and directly generate the object edge map, which serve as the input for template matching.}
\label{fig_template_matching}
\end{figure*}

Given a sequence of measurements, we estimate the object's 3D translation $\mathbf{t}_{wo}$ using a maximum likelihood estimation (MLE) formulation. Assuming a unimodal Gaussian error model, we solve the problem via nonlinear least squares (NLLS) optimization. The optimization is formulated by creating measurement residuals that constrain the object translation $\mathbf{t}_{wo}$ to the object center $\mathbf{u}_k$, its covariance $\boldsymbol{\Sigma}_{\mathbf{u},k}$, and known camera pose $\mathbf{T}_{wc, k}$ at viewpoint $k$,
\begin{equation}
\label{equ_trans_residual}
    \mathbf{r}_k\left(\mathbf{t}_{wo}\right) = \pi\left(\mathbf{T}_{wc, k}^{-1}\mathbf{t}_{wo}\right) - \mathbf{u}_k,
\end{equation}
where $\pi$ denotes the perspective projection function. The full problem then becomes the minimization of the cost function $L$ across all viewpoints,
\begin{equation}
\label{equ_trans_loss}
    L = \sum_k \mathbf{r}_k^T \boldsymbol{\Sigma}_{\mathbf{u},k}^{-1} \mathbf{r}_k,
\end{equation}
where $\boldsymbol{\Sigma}_{\mathbf{u},k}$ is the covariance matrix for the object center $\mathbf{u}_k$. We initialize each object's translation $\mathbf{t}_{wo}$ from two camera viewpoints. For the $k^{th}$ camera viewpoint, $\mathbf{T}_{wc, k}$, we perform object association based on epipolar geometry constraints and the previously estimated translation $\mathbf{t}_{wo, 1:k-1}$ up to viewpoint $k-1$. Detections that are not associated with any existing objects are initialized as new objects.

We solve the NLLS problem (Equation \ref{equ_trans_residual} and \ref{equ_trans_loss}) using an iterative Gauss-Newton procedure:
\begin{equation}
\label{GN_trans}
    \left( \mathbf{J}_{\mathbf{t}_{wo}}^T  \boldsymbol{\Sigma}_{\mathbf{u}}^{-1} \mathbf{J}_{\mathbf{t}_{wo}} \right) \delta \mathbf{t}_{wo} = \mathbf{J}_{\mathbf{t}_{wo}}^T \boldsymbol{\Sigma}_{\mathbf{u}}^{-1} \mathbf{r}\left(\mathbf{t}_{wo}\right),
\end{equation}
where the stacked Jacobian matrix, $\mathbf{J}_{\mathbf{t}_{wo}}$, and the stacked measurement covariance matrix, $\boldsymbol{\Sigma}_{\mathbf{u}}$, are represented by:
\begin{align}
    \mathbf{J}_{\mathbf{t}_{wo}} =
    \begin{bmatrix}    
    \mathbf{J}_{\mathbf{t}_{wo}, \mathbf{u}_{1}} \\
    \vdots \\
    \mathbf{J}_{\mathbf{t}_{wo}, \mathbf{u}_{K}}
    \end{bmatrix}
    \:,\:
    \boldsymbol{\Sigma}_{\mathbf{u}} =
    \begin{bmatrix}
    \boldsymbol{\Sigma}_{\mathbf{u}_{1}} & {} & {}\\
    {} & \ddots & {}\\
    {} & {} & \boldsymbol{\Sigma}_{\mathbf{u}_{K}}\\
    \end{bmatrix}  
\end{align}
The row-blocks, $\mathbf{J}_{\mathbf{t}_{wo}, \mathbf{u}_{k}}$, and $\boldsymbol{\Sigma}_{\mathbf{u}_k}$ correspond to the Jacobian matrix and measurement covariance matrix for the $k^{th}$ viewpoint. The per-frame measurement uncertainty, $\boldsymbol{\Sigma}_{\mathbf{u}_k}$, is obtained from our MEC-Net (upper part of Figure~\ref{fig_pipeline_rgb}). The Jacobian matrix, $\mathbf{J}_{\mathbf{t}_{wo}, \mathbf{u}_{k}}$, is computed as:
\begin{equation}
\label{equ4}
    \mathbf{J}_{\mathbf{t}_{wo}, \mathbf{u}_{k}}=\frac{\partial \mathbf{u}_k}{\partial \mathbf{t}_{wo}}=\frac{\partial \mathbf{u}_k}{\partial \mathbf{t}_{co, k}}\frac{\partial \mathbf{t}_{co, k}}{\partial \mathbf{t}_{wo}},
\end{equation}
where $\mathbf{t}_{co, k}$ is the 3D translation from the object model origin to the camera optical center at the $k^{th}$ viewpoint.

\subsection{3D Orientation Estimation}
\label{Sec:rotation}
The procedure for estimating the object orientation $\mathbf{R}_{wo}$ is shown in the lower part of Figure~\ref{fig_pipeline_rgb}. Given the per-frame edge map from MEC-Net, we first adopt a template matching (TM)-based approach, LINE-2D~\citep{hinterstoisser2011gradient}, to obtain the per-frame orientation measurement $\mathbf{R}_{co, k}$. Measurements from multiple viewpoints are then integrated into an optimization scheme. We handle the rotational symmetries explicitly using the object CAD model. To counteract the measurement uncertainties (e.g., from appearance ambiguities), a max-mixture formulation~\citep{olson2013inference,fu2021multi} is employed to recover a globally consistent set of object orientation estimates. Note that the acquisition of the orientation measurement $\mathbf{R}_{co, k}$ is not limited to the LINE-2D~\citep{hinterstoisser2011gradient} or TM-based approaches and can be superseded by other holistic methods~\citep{liu2012fast,imperoli2015d,kehl2017ssd,sundermeyer2018implicit}.

\vspace{0.8\baselineskip}
\subsubsection{Per-Frame Orientation Measurement}
\hfill\\
\vskip\baselineskip
\vspace{-1.5\baselineskip}
\label{sec_rot_meas}
The process of acquiring the per-frame object orientation measurement, $\mathbf{R}_{co, k}$, is illustrated in Figure~\ref{fig_template_matching}. This process is based on a template-matching (TM)-based approach, LINE-2D~\citep{hinterstoisser2011gradient}. The original LINE-2D method estimates the object's 3D orientation by matching templates derived from the object’s 3D model. In the offline training stage, LINE-2D renders object templates from a view sphere, with each template represented as a set of sampled edge points (shown in the upper part of Figure~\ref{fig_template_matching}). At run-time, it first extracts the edge pixels (e.g., using a Sobel filter) from the input RGB image and utilizes the gradient response to find the best matched template, determining the object orientation. A confidence score is computed based on the quality of the match. However, the template matching-based approach suffers from scale change and occlusion problems. Additionally, specular reflections on shiny surfaces can introduce false edges, leading to incorrect matches. To address these issues, we propose two major modifications to the LINE-2D algorithm.

\textbf{Predicting object edge map}. To bridge the gap between rendered templates and RoI images, and to reduce the impact of spurious edges, we leverage our MEC-Net to directly generate the object's edge map. As illustrated in Figure~\ref{fig_pipeline_rgb}, we extend our previous approach~\citep{yang20236d} by adding an extra network head specifically for estimating the object's 2D edge map. To handle partial occlusion, which is common in real-world scene, we incorporate occlusion augmentation in our training data, similar to the approach used in AAE~\citep{sundermeyer2018implicit}. During training, we treat the edge map as a binary classification task and minimize the cross-entropy loss. At inference, we apply the sigmoid function to map edge pixel values to the range of $[0,1]$, with higher values indicating greater confidence that a pixel belongs to the object edge.

\begin{figure}[t]
\centering
\begin{subfigure}{0.48\textwidth}
  \includegraphics[width=\linewidth]{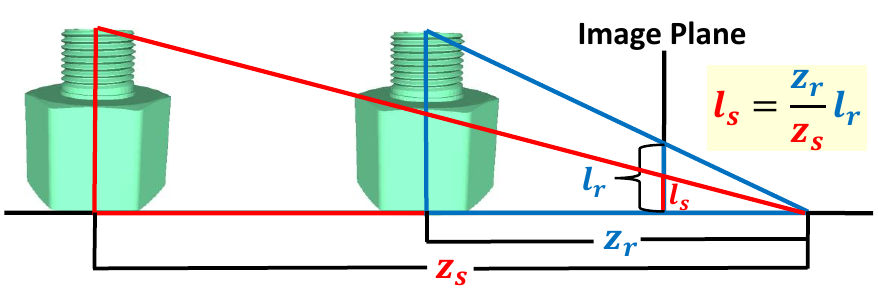}
  \caption{}
  \label{fig_resize_a}
\end{subfigure}
\begin{subfigure}{0.235\textwidth}
    {\includegraphics[width=\linewidth]{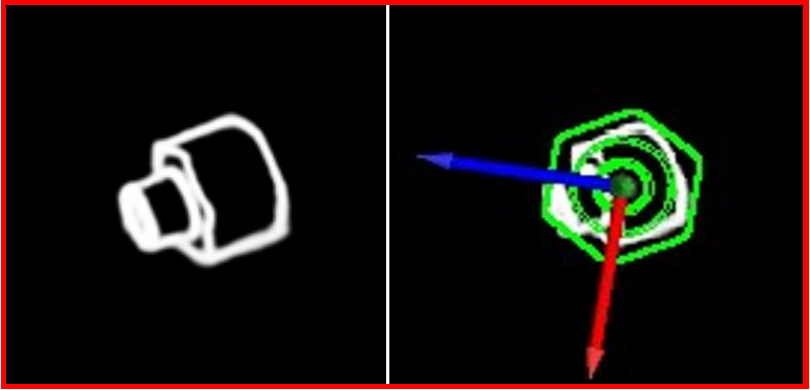}}    
    \caption{}
    \label{fig_resize_b}
\end{subfigure}
\begin{subfigure}{0.235\textwidth}
    {\includegraphics[width=\linewidth]{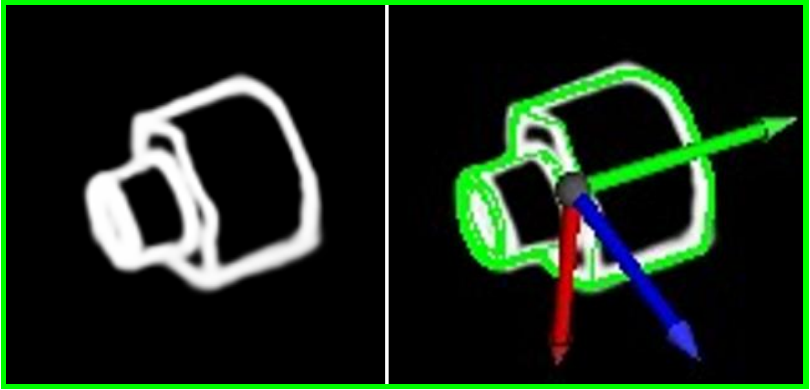}}    
    \caption{}
    \label{fig_resize_c}
\end{subfigure}
\caption{(a). The inference of object size $l_s$ from its projective ratio. (b) Incorrect object orientation estimates on the original edge map due to scale changes. (c) re-cropped object RoI using the object translation estimate, , resulting in correct orientation estimation.}
\label{fig_resize}
\end{figure}

\textbf{Handling object scale change}. To address the scale change issue, the original LINE-2D generates object templates at multiple distances and scales, which increases run-time complexity. In contrast, our approach fixes the 3D translation to a canonical centroid distance, 
\begin{equation}
    \mathbf{t}_{r} = \left[0, 0, z_r\right].
\end{equation}
At run-time, given the 3D translation estimate, $\mathbf{t}_{co} = \left[x_s, y_s, z_s\right]$, from object origin to camera center (obtained from $\mathbf{t}_{wo}$ and camera pose, $\mathbf{T}_{wc}$), we re-crop the edge map RoI from the image. The RoI size $l_s$ is determined by:
\begin{equation}
    l_s = \frac{z_r}{z_s}l_r,
\end{equation}
where $l_r$ and $z_r$ represent the RoI size and canonical distance during training, respectively. This process is illustrated in Figure~\ref{fig_resize_a}. When the translation estimate, $\mathbf{t}_{co}$, is accurate, the resized edge map ROI will have the same size with the rendered object template at the canonical distance. Note that the RoI is square here and independent of the object's orientation. Finally, the per-frame measurement of the object orientation, $\mathbf{R}_{co, k}$, is obtained by feeding the resized RoI into the LINE-2D orientation estimator. As shown in Figure~\ref{fig_resize_b} and \ref{fig_resize_c}, compared to using the original edge map, this resizing step yields a correct orientation estimate.

\vspace{0.8\baselineskip}
\subsubsection{Optimization formulation}
\hfill\\
\vskip\baselineskip
\vspace{-1.5\baselineskip}
Given the multi-view orientation measurements, $\mathbf{R}_{co, k}$, we aim to estimate the object's 3D orientation in the global world frame. Generally, estimating the object's 3D orientation from a sequence of such measurements can also be formulated as an MLE problem:
\begin{align}
    \hat{\mathbf{X}} &= \argmax_{\mathbf{X}} \prod_k p(\mathbf{z}_k|\mathbf{X}),
\end{align}
where $\mathbf{X}$ denotes the object 3D orientation, $\mathbf{R}_{wo}$, to be estimated. The measurement $\mathbf{z}_k$ refers to the object's orientation with respect to the camera coordinate, $\mathbf{R}_{co, k}$, obtained in Section~\ref{sec_rot_meas}. The measurement model is a function of the camera pose (orientation part), $\mathbf{R}_{wc, k}$, and the object's orientation, $\mathbf{R}_{wo}$, in the world frame:
\begin{align}
    h\left(\mathbf{R}_{wo}, \mathbf{R}_{wc, k}\right) = \mathbf{R}_{wc, k}^{-1} \mathbf{R}_{wo}.
\end{align}

We formulate the optimization problem by creating the residual between the object orientation, $\mathbf{R}_{wo}$, and the per-frame measurement, $\mathbf{R}_{co, k}$:
\begin{align}
    \label{rot_residual}
    \mathbf{r}_k \left( \mathbf{R}_{wo} \right)&=
    \log\left({\mathbf{R}_{co, k} h\left(\mathbf{R}_{wo}, \mathbf{R}_{wc, k} \right)^{-1}}\right)^{\vee},
\end{align}
where $\mathbf{r}_k \left( \mathbf{R}_{wo} \right)$ is expressed in Lie algebra ${\mathfrak{so}(3)}$. To account for rotational symmetries, we explicitly consider them alongside the measurement $\mathbf{R}_{co, k}$ in Equation \ref{rot_residual}. Generally, when an object has symmetry, there exists a set of orientations that leave the object's appearance unchanged:
\begin{align}
    \label{rot_symmetry}
    \mathbf{S}\left(\mathbf{R}_{co}\right)&=
    \Bigl\{\mathbf{R}^{\prime}_{co} \in \mathbb{SO}(3)  \:\:s.t\:\: \forall \: \mathcal{G}\bigl(\mathbf{R}_{co}\bigl) = \mathcal{G}\bigl(\mathbf{R}^{\prime}_{co}\bigl) \Bigl\},
\end{align}
where $\mathcal{G}\bigl(\mathbf{R}_{co}\bigl)$ is the rendered image of the object under orientation $\mathbf{R}_{co}$ (assuming the same object translation). We can update the measurement $\mathbf{R}_{co, k}$ in Equation \ref{rot_residual} to $\bar{\mathbf{R}}_{co,k}$:
\begin{align}
\label{equ6}
    \bar{\mathbf{R}}_{co,k}
    &= \argmin_{\mathbf{R}^{\prime}_{co,k} \in \mathbf{S}\left(\mathbf{R}_{co, k}\right)} \: \normx{
    \log\left({\left(\mathbf{R}^{\prime}_{co,k}\right) h\left(\mathbf{R}_{wo}, \mathbf{R}_{wc, k}\right)^{-1}}\right)^{\vee}},
\end{align}
where $\normx{\cdot}$ denotes the absolute angle of a 3D rotation vector $\boldsymbol{\phi}$, and $\bar{\mathbf{R}}_{co,k}$ is the updated orientation measurement that has the minimal loss relative to $\mathbf{R}_{wo}$.

\vspace{0.8\baselineskip}
\subsubsection{Measurement ambiguities}
\hfill\\
\vskip\baselineskip
\vspace{-1.5\baselineskip}
Due to the complex uncertainties, such unimodal estimates are insufficient to fully capture the uncertainty associated with the object orientation. To this end, we now consider the sum-mixture of Gaussians as the likelihood function:
\begin{align}
\label{equ7}
    p(\bar{\mathbf{z}}_k|\mathbf{X}) = \sum_{i=1}^N w_i \mathcal{N}\left( \boldsymbol{\mu}_i, \boldsymbol{\Sigma}_i \right),
\end{align}
where $\bar{\mathbf{z}}_k$ is the updated measurement (using Equation~\ref{equ6}), and each $\mathcal{N} \left( \boldsymbol{\mu}_i, \boldsymbol{\Sigma}_i \right)$ represents a distinct Gaussian distribution, with $w_i$ being the weight for component $i$. A challenge with the sum-mixture model is that the MLE solution becomes more complex and falls outside the support of common NLLS optimization approaches. To address this, we consider the max-marginal and solve the optimization problem using the following max-mixture formulation~\citep{olson2013inference}:
\begin{align}
\label{max_mix_equ}
    p(\bar{\mathbf{z}}_k|\mathbf{X}) &= \max_{i=1:N} w_i \mathcal{N}\left( \boldsymbol{\mu}_i, \boldsymbol{\Sigma}_i \right).
\end{align}
The max operator acts as a selector, reducing the problem to a common NLLS optimization. It’s important to note that the max-mixture does not make a permanent selection. During each iteration of the optimization, only one of the Gaussian components is selected and optimized. Specifically, given a new orientation measurement $\bar{\mathbf{R}}_{co,k}$ at frame $k$, we actually evaluate each Gaussian component in Equation \ref{max_mix_equ} by computing the absolute orientation angle error, $\boldsymbol{\theta}_{k,i}$, between $\bar{\mathbf{R}}_{co,k}$ and $h\left(\mathbf{R}_{wo,i}, \mathbf{R}_{wc, k}\right)$,
\begin{align}
    \boldsymbol{\theta}_{k,i}
    &= \normx{
    \log\left(\bar{\mathbf{R}}_{co,k} h\left(\mathbf{R}_{wo}, \mathbf{R}_{wc, k}\right)^{-1}\right)^{\vee}},
\end{align}
and select the component with the minimal angle error. To reduce the impact of outliers, the selected Gaussian component will only accept an orientation measurement if the orientation angle error $\boldsymbol{\theta}_{k,i}$ is below a pre-defined threshold ($30^{\circ}$ in our implementation). We initialize the Gaussian-mixture model with a single component using the first orientation measurement. As more measurements arrive, each new measurement $\bar{\mathbf{R}}_{co,k}$ is evaluated against the existing components. If the measurement $\bar{\mathbf{R}}_{co,k}$ is not accepted by any Gaussian component, it will be treated as a new component and added to the current Gaussian-mixture model. 

For each Gaussian component, we optimize the object orientation $\mathbf{R}_{wo}$ by constructing the following residual $\mathbf{r}\left(\mathbf{R}_{wo}\right)$ at frame $k$:
\begin{equation}
    \mathbf{r}_k\left(\mathbf{R}_{wo}\right) = 
    \log\left(\bar{\mathbf{R}}_{co,k}^{-1} \mathbf{R}_{wc, k}^{-1} \mathbf{R}_{wo}\right)^{\vee}.
\end{equation}

\begin{figure*}[t]
\centering
\begin{subfigure}{.96\textwidth}
  \includegraphics[width=\linewidth]{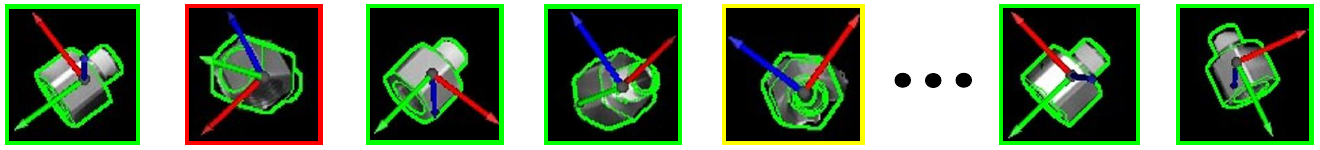}
  \caption{}
\label{fig_maxmix_a}
\end{subfigure}
\begin{subfigure}{.32\textwidth}
  \includegraphics[width=\linewidth]{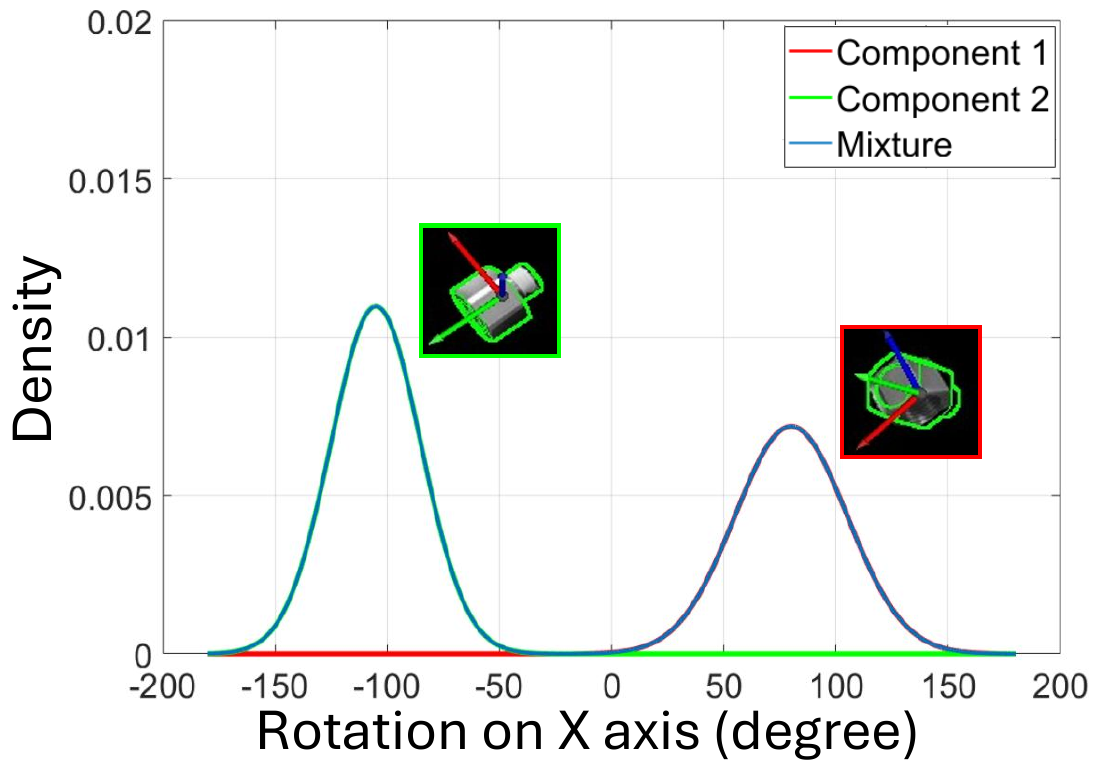}  
  \caption{}
  \label{fig_maxmix_b}
\end{subfigure}
\begin{subfigure}{.32\textwidth}
  \includegraphics[width=\linewidth]{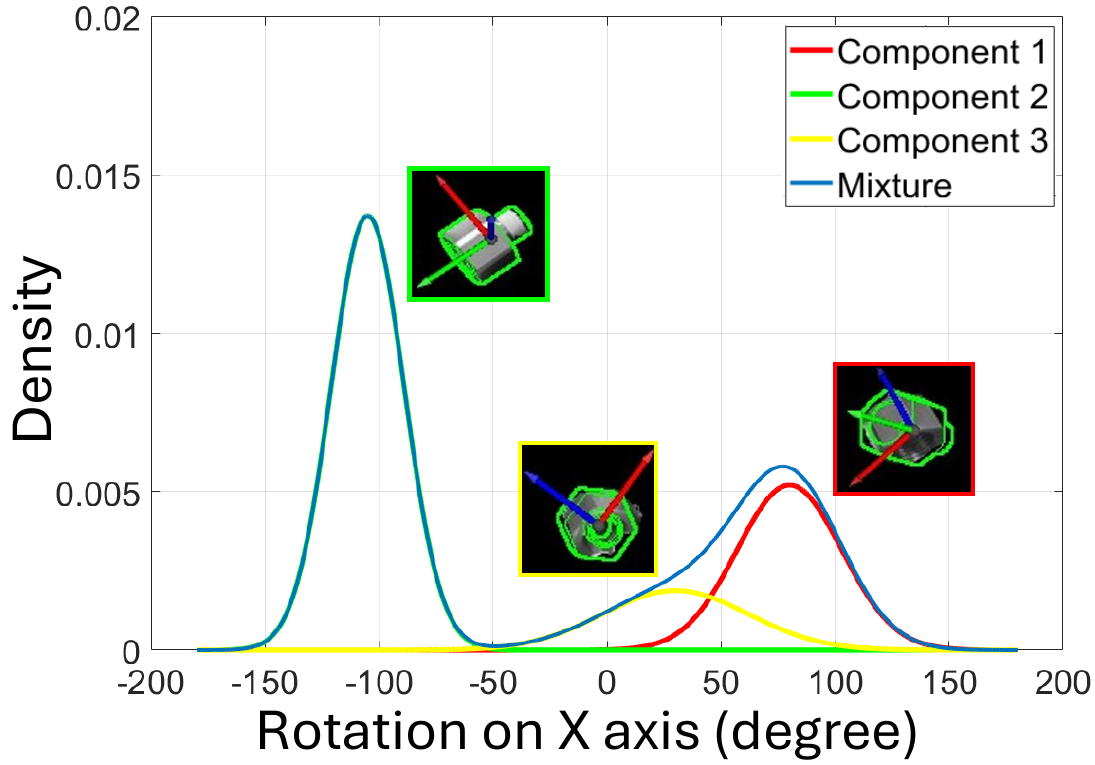} 
  \caption{}
  \label{fig_maxmix_c}
\end{subfigure}
\begin{subfigure}{.32\textwidth}
  \includegraphics[width=\linewidth]{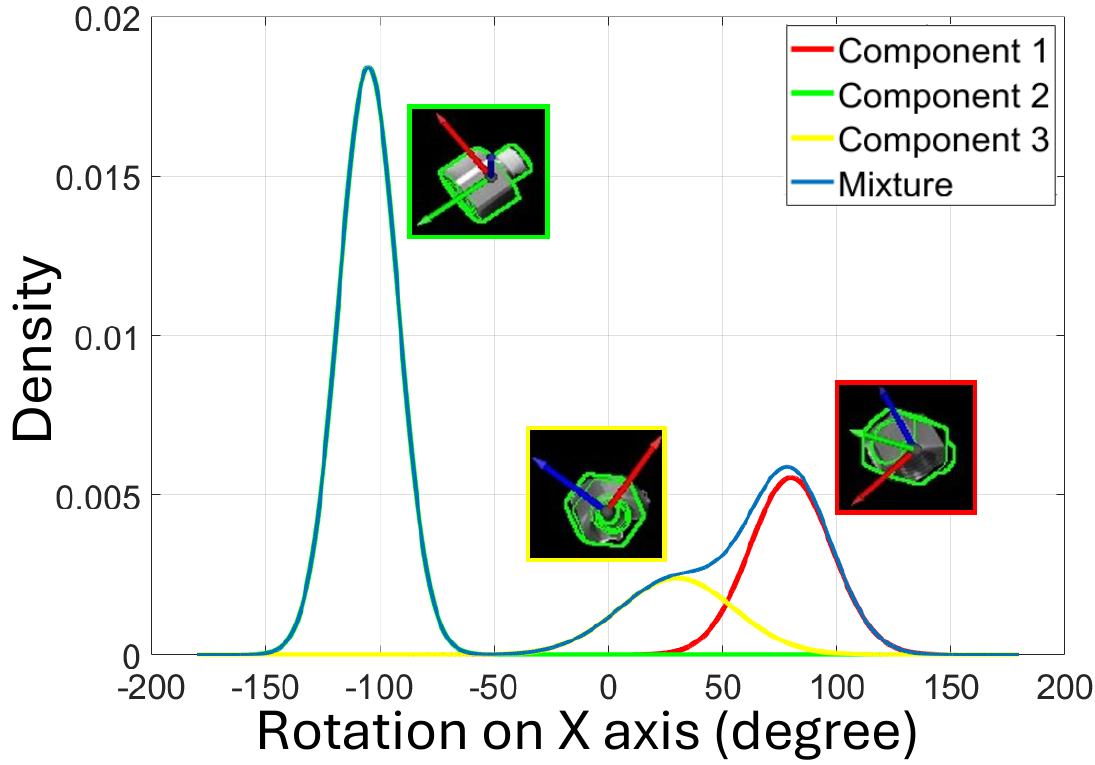} 
  \caption{}
  \label{fig_maxmix_d}
\end{subfigure}
\caption[Max-mixtures for processing the object orientation measurements.]{Max-mixtures for processing the object orientation measurements. Note that we show the distribution only on one axis for demonstration purposes. (a) Acquired orientation measurements from different viewpoints. (b) Mixture distribution after two viewpoints. (c) Mixture distribution after five viewpoints. (d) Mixture distribution after eight viewpoints.}
\label{fig_maxmix}
\end{figure*}

We perform the optimization in the tangent space ${\mathfrak{so}(3)}$. Following the standard Lie algebra derivation method, we apply a left disturbance $\Delta \boldsymbol{\phi}$ to $\mathbf{R}_{wo}$, and the residual error becomes:
\begin{align}
\label{equ_derivation}
    \hat{\mathbf{r}}_k &= \ln{\left( \bar{\mathbf{R}}_{co,k}^{-1} \mathbf{R}_{wc, k}^{-1} \exp{\left(\Delta \boldsymbol{\phi}\right)^{\wedge}} \mathbf{R}_{wo} \right)}^{\vee} \\
    &= \ln{\left( \bar{\mathbf{R}}_{co,k}^{-1} \mathbf{R}_{wc, k}^{-1} \mathbf{R}_{wo} \exp{\left(\mathbf{R}_{wo}^{-1} \Delta \boldsymbol{\phi}\right)^{\wedge}} \right)}^{\vee} \\
    &\approx \ln{\left( \bar{\mathbf{R}}_{co,k}^{-1} \mathbf{R}_{wc, k}^{-1} \mathbf{R}_{wo} \left(1 + \left(\mathbf{R}_{wo}^{-1} \Delta \boldsymbol{\phi}\right)^{\wedge}\right) \right)}^{\vee} \\
    &= \ln{\left( \exp{\left( \mathbf{r}_k \right)}^{\wedge} + \bar{\mathbf{R}}_{co,k}^{-1} \mathbf{R}_{wc, k}^{-1} \mathbf{R}_{wo} \left(\mathbf{R}_{wo}^{-1} \Delta \boldsymbol{\phi}\right)^{\wedge} \right)}^{\vee} \\
    &= \mathbf{r}_k + \ln{\left( \exp{\left( \mathbf{r}_k \right)^{\wedge}} \exp{\left(\mathbf{R}_{wo}^{-1} \Delta \boldsymbol{\phi}\right)} \right)}\\
    &= \mathbf{r}_k+\frac{\partial \mathbf{r}_k}{\partial \Delta \boldsymbol{\phi}} \Delta \boldsymbol{\phi}\\
    &= \mathbf{r}_k+\mathbf{J}_{\boldsymbol{\phi}_{wo,k}} \Delta \boldsymbol{\phi},
\end{align}

where
\begin{align}
\label{equ_jac_orien}
    \mathbf{J}_{\boldsymbol{\phi}_{wo,k}} &= \mathtt{\mathbf{J}_r^{-1}}\left(\mathbf{r}_k\right)\left(\mathbf{R}_{wo}^{-1}\right)\\
    &\approx \mathtt{\mathbf{I}}\left(\mathbf{r}_k\right)\left(\mathbf{R}_{wo}^{-1}\right),
\end{align}
where $\mathtt{\mathbf{J}_r}$ is the right Jacobian of $\mathbb{SO}(3)$, which can be approximated as the identity matrix when the errors are small. The final Jacobian, $\mathbf{J}_{\boldsymbol{\phi}_{wo,k}}$, for each orientation measurement is a $3 \times 3$ matrix.

Similar to the 3D translation approach (Equation~\ref{GN_trans}), we optimize the object orientation, $\mathbf{R}_{wo}$, using the Gauss-Newton solver:
\begin{equation}
\label{equ11}
    \left( \mathbf{J}_{\boldsymbol{\phi}_{wo}}^T  \boldsymbol{\Lambda}_{\boldsymbol{\phi}} \:\mathbf{J}_{\boldsymbol{\phi}_{wo}} \right) \delta \boldsymbol{\phi}_{wo} = \mathbf{J}_{\boldsymbol{\phi}_{wo}}^T \boldsymbol{\Lambda}_{\boldsymbol{\phi}} \:\mathbf{r}\left(\mathbf{R}_{wo}\right),
\end{equation}
where $\mathbf{r}\left(\mathbf{R}_{wo}\right)$ is the stacked rotation residual vector across multiple viewpoints. The stacked Jacobian matrix, $\mathbf{J}_{\boldsymbol{\phi}_{wo}}$, and the stacked measurement weight matrix, $\boldsymbol{\Lambda}_{\boldsymbol{\phi}}$, are given by:
\begin{align}
    \mathbf{J}_{\boldsymbol{\phi}_{wo}} =
    \begin{bmatrix}
    \mathbf{J}_{\boldsymbol{\phi}_{wo,1}} \\
    \vdots \\
    \mathbf{J}_{\boldsymbol{\phi}_{wo,K}}
    \end{bmatrix}
    \:,\:
    \boldsymbol{\Lambda}_{\boldsymbol{\phi}} =
    \begin{bmatrix}
    \boldsymbol{\Lambda}_{\boldsymbol{\phi},1} & {} & {}\\
    {} & \ddots & {}\\
    {} & {} & \boldsymbol{\Lambda}_{\boldsymbol{\phi},K}\\
    \end{bmatrix}  
\end{align}
The row-blocks, $\mathbf{J}_{\boldsymbol{\phi}_{wo,k}}$, and $\boldsymbol{\Lambda}_{\boldsymbol{\phi},k}$ correspond to the Jacobian matrix and the inverse of the measurement covariance matrix for the $k^{th}$ viewpoint. The per-frame Jacobian matrix is obtained from Equation~\ref{equ_jac_orien}. For the weight matrix $\boldsymbol{\Lambda}_{\boldsymbol{\phi},k}$, we approximate it by placing the LINE-2D confidence score on its diagonal elements. 

To compute the weight, $w_i$, for each Gaussian component, we accumulate the LINE-2D confidence score, $c_i$, from the orientation measurements within each component across the viewpoints. The weight can be approximated as: 
\begin{equation}
    w_i = \frac{{c}_i}{\sum_i {c}_i}.
\end{equation}
This processing is illustrated in Figure \ref{fig_maxmix}. Given the measurements from two viewpoints, the object orientation distribution $P\left(\mathbf{R}_{wo}\right)$ is represented with two Gaussian components (green and red) with similar weights. When additional viewpoints are incorporated, a third component (yellow) is added. As more orientation measurements are received (i.e. after eight viewpoints), the weight of the correct component (green) increases, surpassing the other hypotheses.

\section{Active Pose Estimation using Next-Best-View}
\label{sec4}
In Section~\ref{sec_pose}, we solve the multi-view object pose estimation problem using a two-step optimization formulation. However, the accuracy of the estimated object pose heavily depends on the collected RGB measurements from the selected camera viewpoints. Moreover, in many real-world applications, capturing a large number of viewpoints is impractical. To overcome this limitation, we introduce an active object pose estimation process. This approach not only estimates the uncertainty of the object pose but also predicts the next-best-view to minimize that uncertainty.

\subsection{Initialization and Uncertainty Estimation}
We initialize our active object pose estimation process with a collection of measurement sets, $\mathbf{Z}_{1:K}$, from $K$ camera viewpoints and perform iterative optimization to estimate the object’s 6D pose. To bootstrap the system, at least $K=2$ viewpoints are required. As described in Section~\ref{sec_pose}, we decompose the full 6D object pose into 3D translation, $\mathbf{t}_{wo}$, and 3D orientation, $\mathbf{R}_{wo}$. As a result, we compute their uncertainties independently.

\vspace{0.8\baselineskip}
\subsubsection{3D Translation}
\begin{figure*}[t]
\centering
\begin{subfigure}{0.9\textwidth}
    \includegraphics[width=\linewidth]{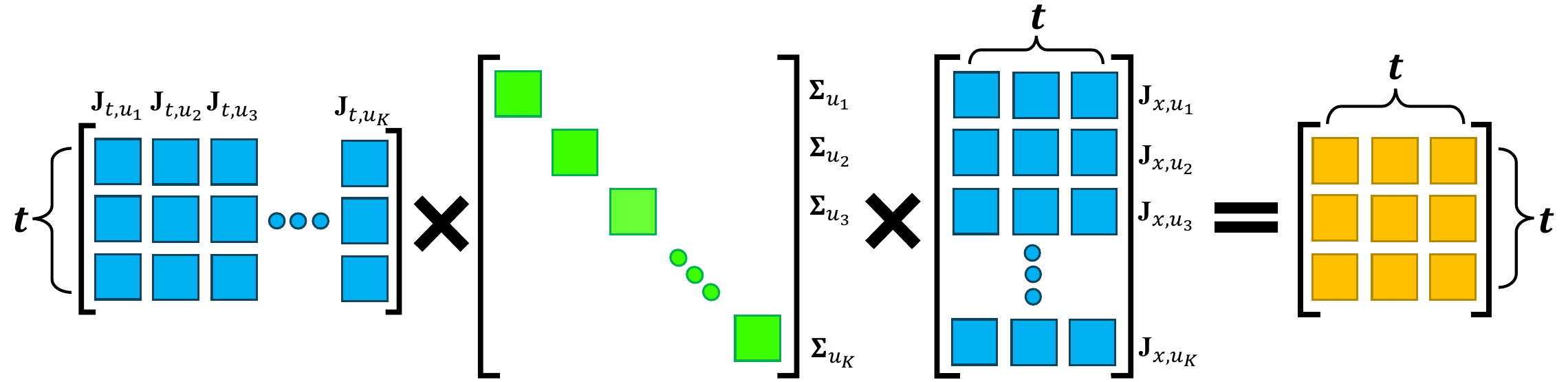}
    \caption{The covariance matrix construction from collected measurements.}
    \label{cov_trans_ini}
\end{subfigure}
\begin{subfigure}{0.9\textwidth}
    \includegraphics[width=\linewidth]{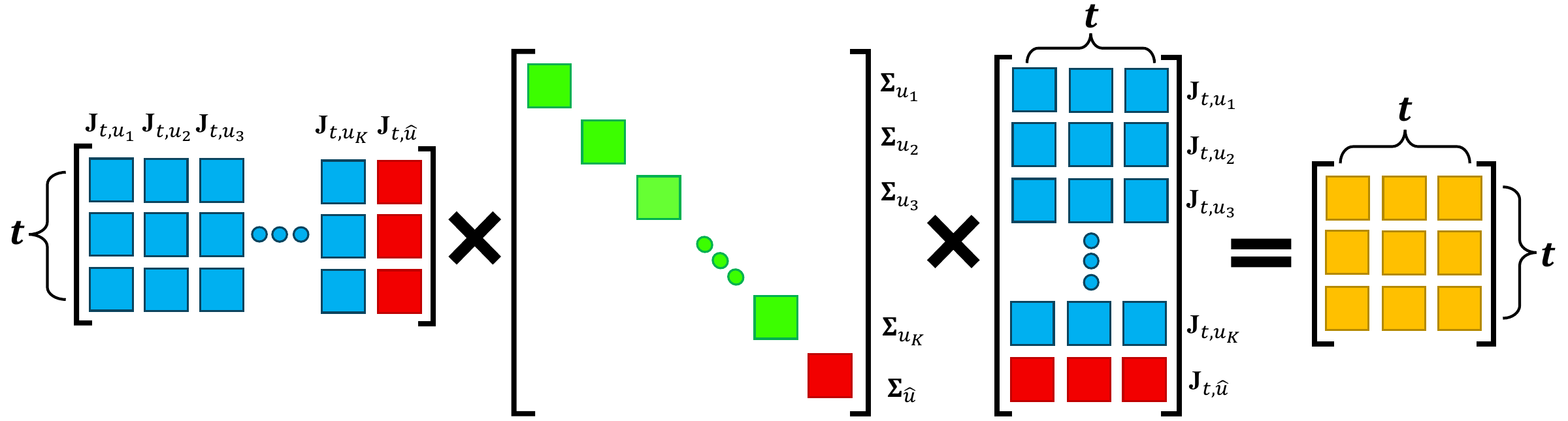}
    \caption{The covariance matrix construction when predicting the NBV.}
    \label{cov_trans_nbv}
\end{subfigure}
\caption{Visualization of the covariance matrix construction for object 3D translation. (a) The construction from collected measurements. (b) The construction when predicting NBV.}
\end{figure*}

\hfill\\
\vskip\baselineskip
\vspace{-1.5\baselineskip}
As discussed in Section~\ref{Sec:translation}, we assume that the object’s translation, $\mathbf{t}_{wo}$, follows a unimodal Gaussian distribution, $\mathcal{N}\left(\mathbf{t}_{wo}|\mathbf{\mu},\boldsymbol{\Sigma}\right)$. The translation is estimated via Equation~\ref{GN_trans}, using the stacked Jacobian, $\mathbf{J}_{\mathbf{t}_{wo}, \mathbf{u}_{1:K}}$, and the stacked measurement uncertainties, $\boldsymbol{\Sigma}_{\mathbf{u}_{1:K}}$:
\begin{align}
    \mathbf{J}_{\mathbf{t}_{wo}, \mathbf{u}_{1:K}} =
    \begin{bmatrix}
    \mathbf{J}_{\mathbf{t}_{wo}, \mathbf{u}_{1}} \\
    \vdots \\
    \mathbf{J}_{\mathbf{t}_{wo}, \mathbf{u}_{K}}
    \end{bmatrix}
    ,
    \boldsymbol{\Sigma}_{\mathbf{u}_{1:K}} =
    \begin{bmatrix}
    \boldsymbol{\Sigma}_{\mathbf{u}_1} & {} & {}\\
    {} & \ddots & {}\\
    {} & {} & \boldsymbol{\Sigma}_{\mathbf{u}_K}\\
    \end{bmatrix},
\end{align}
where $\mathbf{u}_{1:K}$ denotes the object 2D center measurements from $K$ camera viewpoints. As illustrated in Figure~\ref{cov_trans_ini}, we compute the covariance of the object translation, $\boldsymbol{\Sigma}_{\mathbf{t}_{wo},\mathbf{u}_{1:K}}$, through a first-order approximation of the Fisher information matrix (FIM):
\begin{equation}
\label{equ_trans_cov}
\boldsymbol{\Sigma}_{\mathbf{t}_{wo},\mathbf{u}_{1:K}} = {\left( \mathbf{J}_{\mathbf{t}_{wo}, \mathbf{u}_{1:K}}^T \: \boldsymbol{\Sigma}_{\mathbf{u}_{1:K}}^{-1} \: \mathbf{J}_{\mathbf{t}_{wo}, \mathbf{u}_{1:K}} \right)}^{-1}.
\end{equation}
To obtain the entropy from the translation covariance matrix, we employ the differential entropy, $h_e\left( \boldsymbol{\Sigma}_{\mathbf{t}_{wo},\mathbf{u}_{1:K}} \right)$:
\begin{align}
\label{equ_trans_entropy}
h_{\mathbf{t}_{wo}} = h_e\left( \boldsymbol{\Sigma}_{\mathbf{t}_{wo},\mathbf{u}_{1:K}} \right) = \frac{1}{2} \ln{\left(\left(2\pi e\right)^3 \left| \boldsymbol{\Sigma}_{\mathbf{t}_{wo}, \mathbf{u}_{1:K}} \right| \right)},
\end{align}
where $h_{\mathbf{t}_{wo}}$ is expressed in nats. Note that the entropy computation is not restricted to differential entropy and can be replaced by alternative metrics, such as the trace~\citep{costante2018exploiting} or the sum of the eigenvalues of the covariance matrix~\citep{kiciroglu2020activemocap}.

\begin{figure}[t]
\centering
\begin{subfigure}{0.21\textwidth}
  \includegraphics[width=\linewidth]{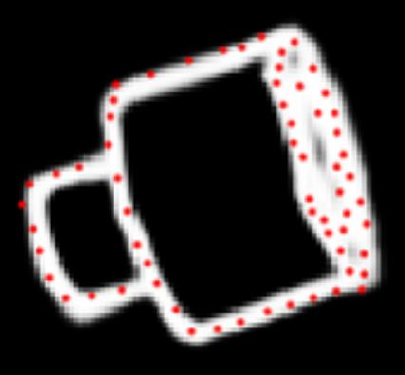}
  \caption{Accurate edge alignment}
\end{subfigure}
\begin{subfigure}{0.21\textwidth}
    \includegraphics[width=\linewidth]{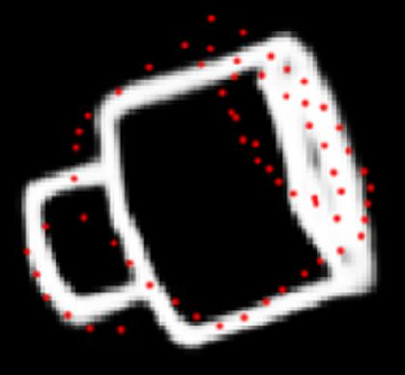}
      \caption{Inaccurate edge alignment.}
\end{subfigure}
\caption{(a) Low orientation uncertainty when edge alignment is accurate (\( h_{\boldsymbol{\phi}_{wo}} = 3.21 \) nats).  
(b) High orientation uncertainty when edge alignment is not accurate (\( h_{\boldsymbol{\phi}_{wo}} = 3.73 \) nats).}
\label{fig_edge_alignment}
\end{figure}

\vspace{0.8\baselineskip}
\subsubsection{3D Orientation}
\hfill\\
\vskip\baselineskip
\vspace{-1.5\baselineskip}
In contrast, the uncertainty calculation for the object orientation is more complex due to the Gaussian mixture formulation, $\sum_{i=1}^{N}w_i \mathcal{N}\left(\boldsymbol{\phi}_{wo}|\boldsymbol{\mu}_i,\boldsymbol{\Sigma}_i\right)$, as shown in Equation~\ref{equ_rot_dist}. While many approaches estimate entropy using sampling methods~\citep{shi2021fast}, which can be computationally expensive, we instead follow the approach from~\citep{eidenberger2010active} and estimate the orientation entropy using an upper bound approximation for the Gaussian mixture distribution~\citep{huber2008entropy}:
\begin{align}
\label{equ_gmm_entropy}
h_{\boldsymbol{\phi}_{wo}} \leq h_{\boldsymbol{\phi}_{wo}}^u = \sum_{i=1}^{N}w_i\left[-\ln{w_i}+\frac{1}{2} \ln{\left(\left(2\pi e\right)^3 \left| \boldsymbol{\Sigma}_i \right| \right)}\right],
\end{align}
where $h_{\boldsymbol{\phi}{wo}}$ is the true entropy of the Gaussian mixture, and $h{\boldsymbol{\phi}_{wo}}^u$ is the upper bound approximation. To estimate the entropy for each individual Gaussian covariance, $\boldsymbol{\Sigma}_i$, we re-project the 3D edge points (from model templates) into the image space and evaluate the alignment quality between the projected points and the 2D edge maps from different camera viewpoints. As shown in Figure~\ref{fig_edge_alignment}, the orientation uncertainty is low when the re-projected edge points align well with the 2D edge map, and high when the alignment is poor.

We begin by deriving the Jacobian of the projected edge points and their associated measurement uncertainties, which will later be used to compute the orientation covariance. For a set of $N$ 3D model edge points, we denote their stacked coordinates as $\mathbf{o}_k \in \mathbb{R}^{3N}$. After transforming these points into the $k^{th}$ camera viewpoint, each point is reprojected onto the image plane, producing a corresponding 2D pixel location. By stacking these $N$ projected points, we obtain the measurement vector $\mathbf{m}_k \in \mathbb{R}^{2N}$. The Jacobian, $\mathbf{J}_{\boldsymbol{\phi}_{wo}, \mathbf{m}_{k}}$, is then expressed as:
\begin{equation}
\label{equ_jac_rot_unc}
    \mathbf{J}_{\boldsymbol{\phi}_{wo}, \mathbf{m}_{k}}=\frac{\partial \mathbf{m}_k}{\partial \boldsymbol{\phi}_{wo}}=\frac{\partial \mathbf{m}_k}{\partial \mathbf{p}_{c, k}}\frac{\partial \mathbf{p}_{c, k}}{\partial \mathbf{p}_w} \frac{\partial \mathbf{p}_w}{\partial \boldsymbol{\phi}_{wo}},
\end{equation}
where $\mathbf{p}_{c, k} \in \mathbb{R}^{3N}$ and $\mathbf{p}_w \in \mathbb{R}^{3N}$ denote the object’s 3D edge points in the $k^{th}$ camera frame and the world frame, respectively. Since the measurement vector $\mathbf{m}_k \in \mathbb{R}^{2N}$ and the orientation vector $\boldsymbol{\phi}_{wo} \in \mathfrak{so}(3)$, the Jacobian $\mathbf{J}_{\boldsymbol{\phi}_{wo}, \mathbf{m}_{k}}$ has dimensions $2N \times 3$. Note that the Jacobian, $\mathbf{J}_{\boldsymbol{\phi}_{wo}, \mathbf{m}_{k}}$, is distinct from the Jacobian, $\mathbf{J}_{\boldsymbol{\phi}_{wo}, k}$, in Equation~\ref{equ_jac_orien}. For the associated measurement uncertainties, $\boldsymbol{\Sigma}_{\mathbf{m}_k}$, we approximate them with the inverse of the edge map intensity, placing the values along the diagonal elements and its dimension is $2N \times 2N$. The stacked Jacobian, $\mathbf{J}_{\boldsymbol{\phi}_{wo}, \mathbf{m}_{1:K}}$, and stacked measurement uncertainties, $\boldsymbol{\Sigma}_{\mathbf{m}_{1:K}}$, are given by:
\begin{align}
    \mathbf{J}_{\boldsymbol{\phi}_{wo}, \mathbf{m}_{1:K}} =
    \begin{bmatrix}
    \mathbf{J}_{\boldsymbol{\phi}_{wo}, \mathbf{m}_{1}} \\
    \vdots \\
    \mathbf{J}_{\boldsymbol{\phi}_{wo}, \mathbf{m}_{K}}
    \end{bmatrix}
    ,
    \boldsymbol{\Sigma}_{\mathbf{m}_{1:K}} =
    \begin{bmatrix}
    \boldsymbol{\Sigma}_{\mathbf{m}_1} & {} & {}\\
    {} & \ddots & {}\\
    {} & {} & \boldsymbol{\Sigma}_{\mathbf{m}_K}\\
    \end{bmatrix}.
\end{align}

The covariance matrix, $\boldsymbol{\Sigma}_{\boldsymbol{\phi}_{wo},\mathbf{m}_{1:K}}$ is finally computed using the Fisher information approximation:
\begin{equation}
\label{equ_rot_cov}
\boldsymbol{\Sigma}_{\boldsymbol{\phi}_{wo},\mathbf{m}_{1:K}} = {\left(\mathbf{J}_{\boldsymbol{\phi}_{wo}, \mathbf{m}_{1:K}}^T \: \boldsymbol{\Sigma}_{\mathbf{m}_{1:K}}^{-1} \: \mathbf{J}_{\boldsymbol{\phi}_{wo}, \mathbf{m}_{1:K}} \right)}^{-1}.
\end{equation}

To compute the total entropy over the Gaussian mixture, we apply Equations~\ref{equ_jac_rot_unc}-\ref{equ_rot_cov} to each Gaussian component and substitute the results into Equation~\ref{equ_gmm_entropy}:
\begin{align}
h_{\boldsymbol{\phi}_{wo}}^u &= \sum_{i=1}^{N}w_i\left[-\ln{w_i} + h_e \left( \boldsymbol{\Sigma}_{\boldsymbol{\phi}_{wo}, \mathbf{m}_{1:K},i} \right)\right]\\
&=\sum_{i=1}^{N}w_i\left[-\ln{w_i} + \frac{1}{2} \ln{\left(\left(2\pi e\right)^3 \left| \boldsymbol{\Sigma}_{\boldsymbol{\phi}_{wo}, \mathbf{m}_{1:K},i} \right| \right)}\right],
\end{align}
where $h_{\boldsymbol{\phi}_{wo}}^u$ is expressed in nats. The final entropy of the 6D object pose is given by:
\begin{equation}
h_{6D} = g_t\:h_{\mathbf{t}_{wo}}+ g_{\phi}\:h_{\boldsymbol{\phi}_{wo}}^u,
\end{equation}
where $g_t$ and $g_{\phi}$ are the weights assigned to the translation and orientation entropies, respectively.

\subsection{Next-Best-View Prediction}
\label{NBV}
In our next-best-view setup, we operate with a predefined set of camera viewpoints, $\mathcal{V}$. To improve object pose accuracy, we aim to select the next-best viewpoint $\mathbf{v}^* \in \mathcal{V}$ that minimizes the entropy of the object pose. To deploy this setup on a real robot platform, we could first define the set $\mathcal{V}$, and then map each camera viewpoint to a corresponding robot pose (joint configuration) using the robot’s kinematics.

Suppose we have already collected object center measurements, $\mathbf{u}_{1:K}$, and edge measurements, $\mathbf{m}_{1:K}$, from $K$ different camera viewpoints. For a future camera viewpoint, $\widehat{\mathbf{v}}$, the stacked Jacobian, $\mathbf{J}_{\mathbf{t}_{wo}, \overline{\mathbf{u}}}$, and the stacked measurement uncertainties, $\boldsymbol{\Sigma}_{\mathbf{t}_{wo}, \overline{\mathbf{u}}}$, of the object translation are expressed as follows:
\begin{align}
\label{equ_stack_trans_nbv}
\mathbf{J}_{\mathbf{t}_{wo}, \overline{\mathbf{u}}} &=
\begin{bmatrix}
\mathbf{J}_{\mathbf{t}_{wo}, \mathbf{u}_{1:K}} \\
\mathbf{J}_{\mathbf{t}_{wo}, \widehat{\mathbf{u}}}
\end{bmatrix},    
\\
\boldsymbol{\Sigma}_{\mathbf{t}_{wo}, \overline{\mathbf{u}}} &=
\begin{bmatrix}
\boldsymbol{\Sigma}_{\mathbf{t}_{wo}, \mathbf{u}_{1:K}} & \mathbf{0} \\
\mathbf{0} & \boldsymbol{\Sigma}_{\mathbf{t}_{wo}, \widehat{\mathbf{u}}}
\end{bmatrix}.
\end{align}
where $\overline{\mathbf{u}} = \bigl\{ \mathbf{u}_{1:K}, \widehat{\mathbf{u}} \bigl\}$ includes the set of acquired object center measurements $\mathbf{u}_{1:K}$ from viewpoints $\mathbf{v}_{1:K}$ and the predicted measurement $\widehat{\mathbf{u}}$ for the future viewpoint, $\widehat{\mathbf{v}}$.

For the orientation component, similarly, for each Gaussian component, we define the stacked Jacobian, $\mathbf{J}_{\boldsymbol{\phi}_{wo}, \overline{\mathbf{m}}}$, and the associated measurement uncertainties, $\boldsymbol{\Sigma}_{\boldsymbol{\phi}_{wo}, \overline{\mathbf{m}}}$, for the future viewpoint as follows:

\begin{align}
\mathbf{J}_{\boldsymbol{\phi}_{wo}, \overline{\mathbf{m}}} &=
\begin{bmatrix}
\mathbf{J}_{\boldsymbol{\phi}_{wo}, \mathbf{m}_{1:K}} \\
\mathbf{J}_{\boldsymbol{\phi}_{wo}, \widehat{\mathbf{m}}}
\end{bmatrix},    
\\
\label{equ_stack_rot_nbv}
\boldsymbol{\Sigma}_{\boldsymbol{\phi}_{wo}, \overline{\mathbf{m}}} &=
\begin{bmatrix}
\boldsymbol{\Sigma}_{\boldsymbol{\phi}_{wo}, \mathbf{m}_{1:K}} & \mathbf{0} \\
\mathbf{0} & \boldsymbol{\Sigma}_{\boldsymbol{\phi}_{wo}, \widehat{\mathbf{m}}}
\end{bmatrix}.
\end{align}
Using the Fisher information, we can predict the covariance of the object translation and orientation as:
\begin{equation}
\label{equ_cov_trans_nbv}    \boldsymbol{\Sigma}_{\mathbf{t}_{wo},\overline{\mathbf{u}}} = {\left( \mathbf{J}_{\mathbf{t}_{wo}, \overline{\mathbf{u}}}^T \:\: \boldsymbol{\Sigma}_{\mathbf{t}_{wo}, \overline{\mathbf{u}}}^{-1} \:\: \mathbf{J}_{\mathbf{t}_{wo}, \overline{\boldsymbol{u}}} \right)}^{-1},
\end{equation}
\begin{equation}
\label{equ_cov_rot_nbv}    \boldsymbol{\Sigma}_{\boldsymbol{\phi}_{wo},\overline{\mathbf{m}}} = {\left(\mathbf{J}_{\boldsymbol{\phi}_{wo}, \overline{\mathbf{m}}}^T \:\: \boldsymbol{\Sigma}_{\boldsymbol{\phi}_{wo}, \overline{\mathbf{m}}}^{-1} \:\: \mathbf{J}_{\boldsymbol{\phi}_{wo}, \overline{\mathbf{m}}} \right)}^{-1}.
\end{equation}
We illustrate this process for the translation component in Figure~\ref{cov_trans_nbv}. Note that, in Equations~\ref{equ_stack_trans_nbv} to \ref{equ_stack_rot_nbv}, we compute the Jacobians, $\mathbf{J}_{\mathbf{t}_{wo}, \widehat{\mathbf{u}}}$, $\check{\mathbf{J}}_{\boldsymbol{\phi}_{wo}, \widehat{\mathbf{m}}}$, and measurement uncertainties, $\boldsymbol{\Sigma}_{\mathbf{t}_{wo}, \widehat{\mathbf{u}}}$, $\boldsymbol{\Sigma}_{\boldsymbol{\phi}_{wo}, \widehat{\mathbf{m}}}$, prior to actually moving to the future camera viewpoint $\widehat{\mathbf{v}}$. These Jacobians are computed based on the object pose estimate derived from the measurements $\mathbf{u}_{1:K}$ and $\mathbf{m}_{1:K}$. For measurement uncertainties, we assume that they remain constant across different future viewpoints.

We determine our NBV from the candidate viewpoint set, $\mathcal{V}$, by minimizing the weighted sum of the translation and orientation entropy:
\begin{multline}
\label{equ_IG_max}
\mathbf{v}^{*} = \argmin_{\widehat{\mathbf{v}}} \:\: g_t\:{h_e}\left(\boldsymbol{\Sigma}_{\mathbf{t}_{wo},\overline{\mathbf{u}}}\right) \\
\qquad + g_{\phi}\: \sum_{i=1}^{N}w_i\left[-\ln{w_i} + h_e \left( \boldsymbol{\Sigma}_{\boldsymbol{\phi}_{wo}, \overline{\mathbf{m}},i} \right)\right],
\end{multline}
where $g_t$ and $g_{\phi}$ are the entropy weights for the translation and orientation components, respectively. For the orientation term, this formulation refines accuracy by considering all modes in the orientation distribution, each weighted by its corresponding Gaussian component weight $w_i$. Such a formulation aims to improve the final pose accuracy under the assumption that the initial object pose has no view ambiguity, reducing uncertainty only in such cases.

Once the next-best-view $\mathbf{v}^{*}$ is determined, the camera is moved, and new measurements, $\mathbf{u}^*$, $\mathbf{m}^*$, are collected from the corresponding viewpoint. These new measurements are then appended as follows:
\begin{equation}
\mathbf{u}_{1:K} \cup \mathbf{u}^{*} \rightarrow \mathbf{u}_{1:K+1} \:\:\:,\:\:\: \mathbf{m}_{1:K} \cup \mathbf{m}^{*} \rightarrow \mathbf{m}_{1:K+1}.
\end{equation}
The object translation and orientation are then recomputed, and the NBV selection process is repeated using Equations~(\ref{equ_stack_trans_nbv})–(\ref{equ_IG_max}). This iterative process continues until the predicted entropy falls below a user-defined threshold or until a maximum number of viewpoints has been selected.

\begin{figure*}[t]
\centering
\begin{subfigure}{0.19\textwidth}
  \includegraphics[width=\linewidth]{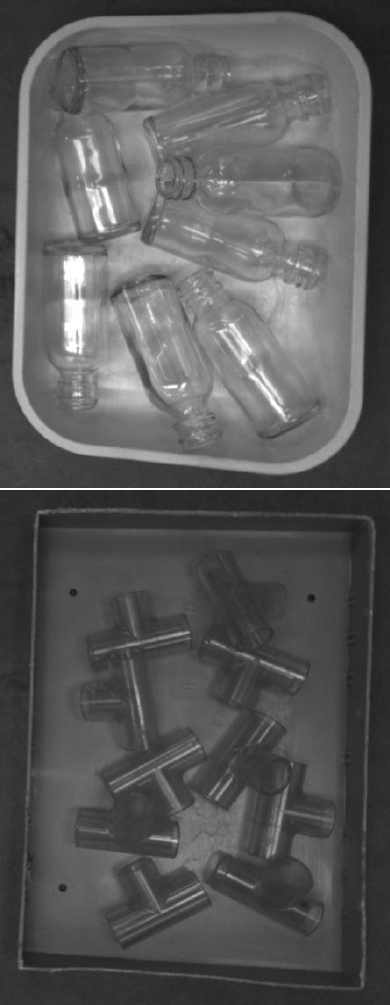}
  \caption{Monochrome images.}
\end{subfigure}
\begin{subfigure}{0.19\textwidth}
  \includegraphics[width=\linewidth]{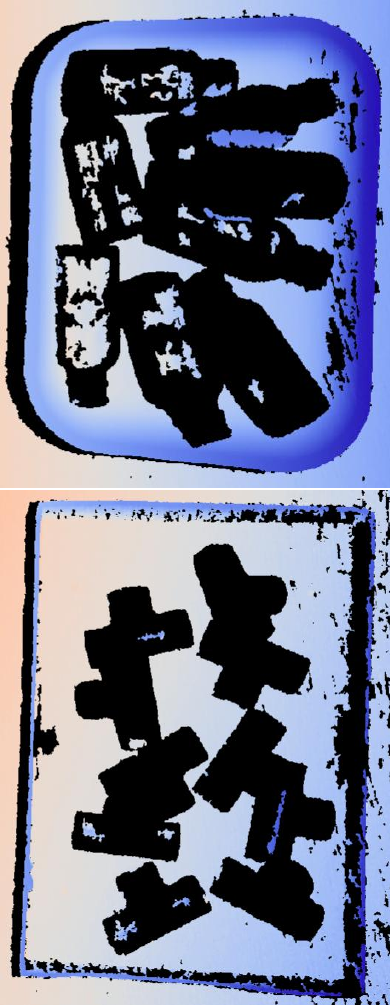}
  \caption{Raw depth maps.}
  \label{fig_TOD_depth}
\end{subfigure}
\begin{subfigure}{0.19\textwidth}
  \includegraphics[width=\linewidth]{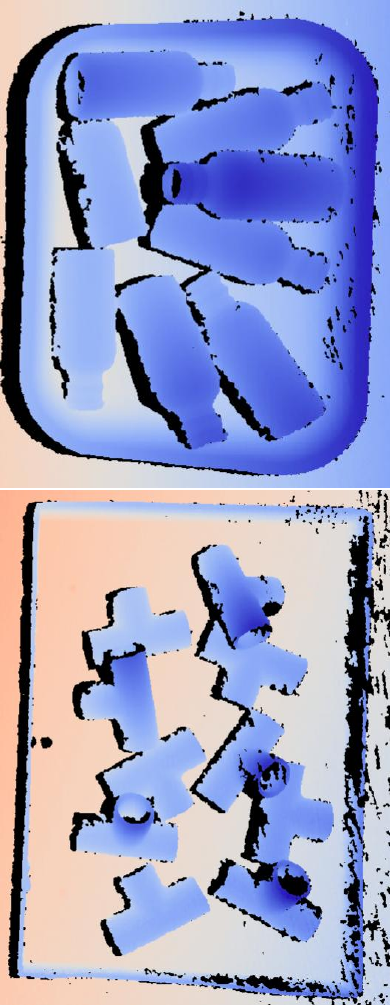}
  \caption{GT depth maps.}
  \label{fig_TOD_gt_depth}
\end{subfigure}
\begin{subfigure}{0.19\textwidth}
  \includegraphics[width=\linewidth]{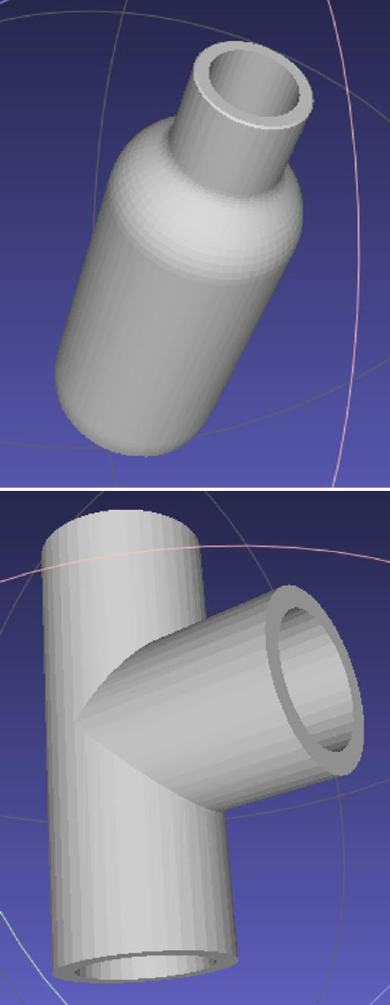}
  \caption{Object CAD models.}
  \label{fig_TOD_cad}
\end{subfigure}
\begin{subfigure}{0.19\textwidth}
  \includegraphics[width=\linewidth]{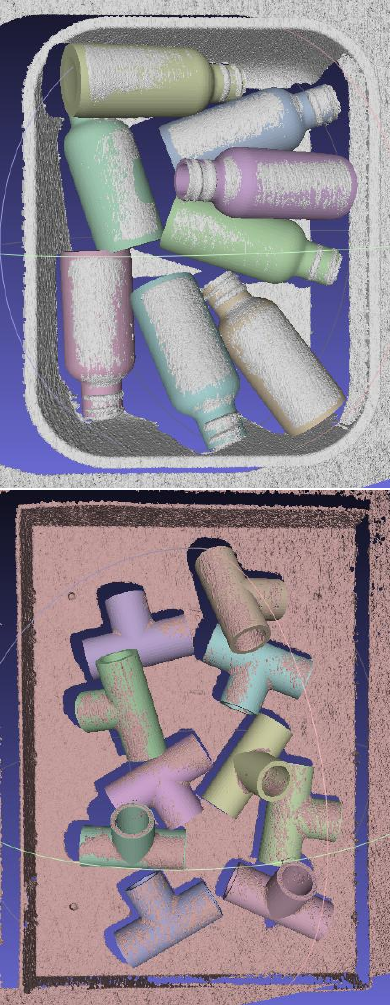}
  \caption{Object GT poses.}
  \label{fig_TOD_gt_poses}
\end{subfigure}
\caption{\textbf{T-ROBI} dataset: (upper) the object ``Bottle'' and (lower) the object ``Pipe Fitting''. (a) Monochrome images. (b) Raw depth maps. (c) Ground truth depth maps. (d) 3D CAD models of the objects. (e) Ground truth 6D object poses.}
\label{fig_TOD}
\end{figure*}

\begin{figure*}[t]
\centering
\begin{subfigure}{0.24\textwidth}
  \includegraphics[width=\linewidth]{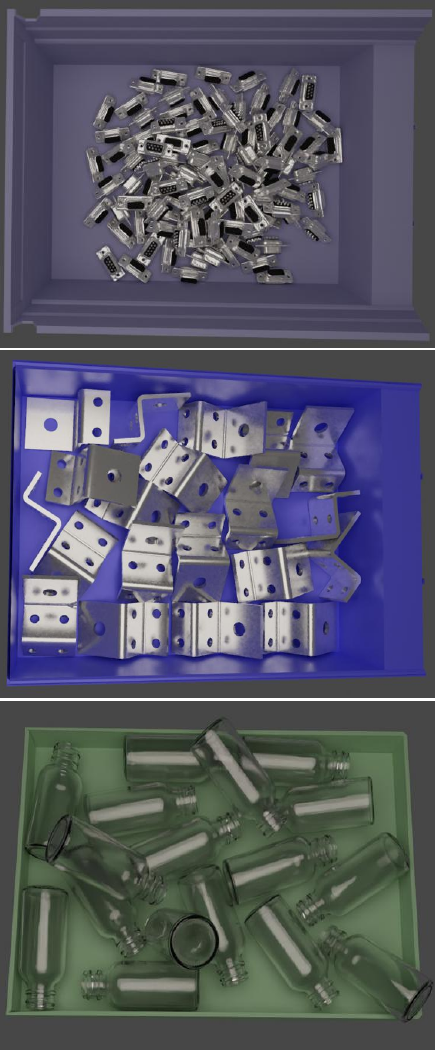}
  \caption{RGB images.}
\end{subfigure}
\begin{subfigure}{0.24\textwidth}
  \includegraphics[width=\linewidth]{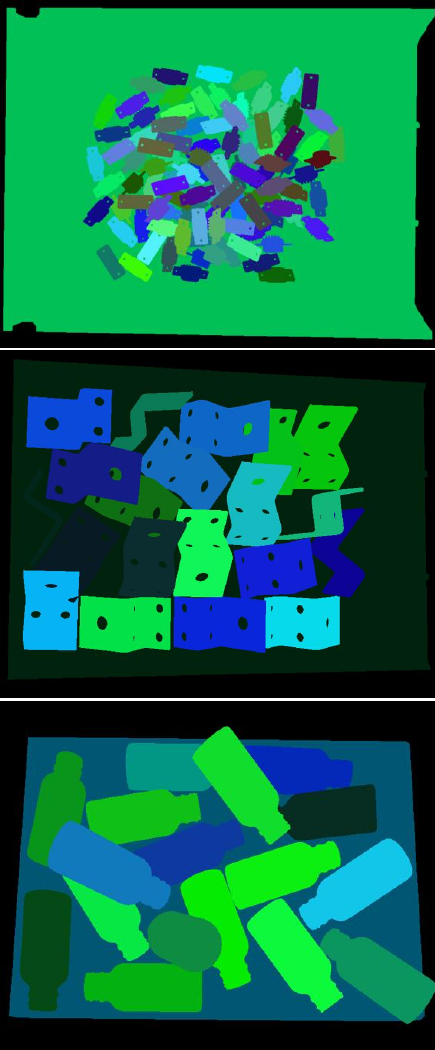}
  \caption{Object masks.}
\end{subfigure}
\begin{subfigure}{0.24\textwidth}
  \includegraphics[width=\linewidth]{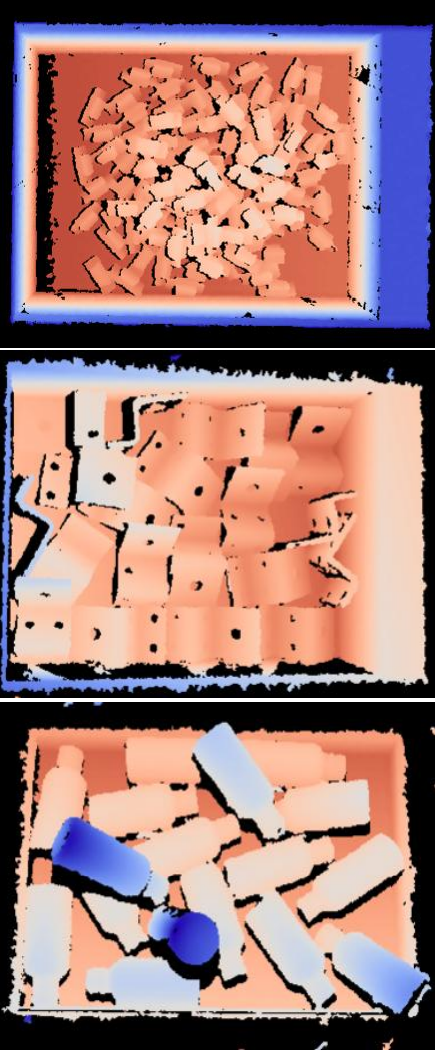}
  \caption{Depth maps.}
  \label{fig_sim_depth}
\end{subfigure}
\begin{subfigure}{0.24\textwidth}
  \includegraphics[width=\linewidth]{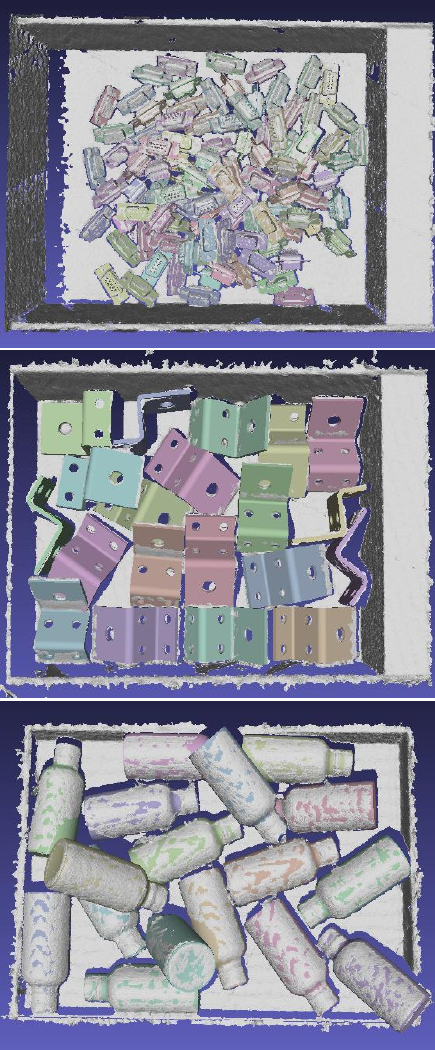}
  \caption{GT object 6D poses.}
\end{subfigure}
\caption{Examples of our generated \textbf{synthetic data} using the Blender rendering software~\citep{blender} with the Bullet physics engine~\citep{coumans2016pybullet}. (a) The RGB images. (b) The object masks. (c) The depth maps. (d) The ground truth 6D object poses. From top to bottom: the object ``D-Sub Connector'', ``Zigzag'' from ROBI dataset~\citep{yang2021robi} and ``Bottle'' from T-ROBI dataset.}
\label{fig_simulation}
\end{figure*}

\section{Experiments}
\label{sec5}
\subsection{Datasets}
\label{sec_dataset}
We evaluate our framework on three challenging real-world datasets: the public ROBI~\citep{yang2021robi} and TOD~\citep{liu2020keypose} datasets, and a new dataset of textureless transparent objects, T-ROBI, which we created for this work. The ROBI dataset contains seven textureless reflective industrial parts placed in complex bin scenarios, recorded from multiple viewpoints using two sensors: a high-end Ensenso camera and a commodity-level RealSense camera. The TOD dataset contains 15 transparent objects across six categories, each captured in isolated settings with diverse backgrounds and multiple RGB-D views per scene.

\textbf{T-ROBI Dataset}. To further validate the effectiveness of our approach, we introduce the T-ROBI (Transparent Reflective Objects in BIns) dataset. This dataset includes two representative components: a ``Bottle'' and a ``Pipe Fitting'', as illustrated in Figure~\ref{fig_TOD}. Unlike other publicly available transparent object datasets~\citep{sajjan2020clear,liu2020keypose,xu2021seeing}, which typically focus on isolated objects, our dataset presents a more challenging scenario. It consists of images containing multiple identical parts randomly stacked within a bin, thereby significantly increasing the difficulty of object pose estimation. For each object, we captured $6$ distinct scenes from $55$ camera viewpoints using the high-end Ensenso N35 camera~\citep{ENSENSO}. For each viewpoint, both monochrome images and depth maps are provided. However, as illustrated in Figure~\ref{fig_TOD_depth}, the transparency of the objects results in significant depth inaccuracies or missing data, making it particularly challenging to label ground truth 6D object poses. To address this, we adopted the ground truth labeling method from the ROBI dataset~\citep{yang2021robi}, utilizing a scanning spray~\citep{AESUB} to capture accurate ground truth depth maps of all bins. The example ground truth depth map, object CAD model, and annotated 6D object poses of the T-ROBI dataset are shown in Figures~\ref{fig_TOD_gt_depth}, \ref{fig_TOD_cad}, and \ref{fig_TOD_gt_poses}, respectively. Upon the publication of this work, we will release a public version of our T-ROBI dataset. This dataset is designed to support 6D pose estimation~\citep{liu2020keypose, chen2023stereopose} as well as depth estimation tasks~\citep{sajjan2020clear, xu2021seeing} for transparent objects in challenging cluttered and occluded bin-picking scenes.

\textbf{Synthetic Dataset}. To facilitate network training, we introduce a large-scale synthetic dataset comprising objects from both the ROBI and T-ROBI datasets, as illustrated in Figure~\ref{fig_simulation}. For each scene, we provide the RGB images, depth maps, object masks, and 6D poses. Our simulation environment is built using the Bullet physics engine~\citep{coumans2016pybullet} in conjunction with Blender software~\citep{blender}. The process begins with importing each object's CAD model into Blender, where we manually specify its color and material properties. After preparing the object, we load it into the simulation and drop it from various positions and orientations within the bin using the Bullet physics engine. This approach allows us to generate a wide variety of object poses, clutter levels, and occlusions. Next, we adjust both the light source and camera pose to different viewpoints above the bin and render the scene using Blender, resulting in high-quality visual representations for our dataset. Finally, we utilize the Ensenso SDK~\citep{ENSENSO} to generate synthetic depth images, as shown in Figure~\ref{fig_sim_depth}. For each object, we produce approximately 6,000 to 13,000 images. We will also release our synthetic dataset upon publication.
\\

\begin{figure*}[t]
\centering
  \parbox{0.97\linewidth}{
    \centering
    \includegraphics[width=\linewidth]{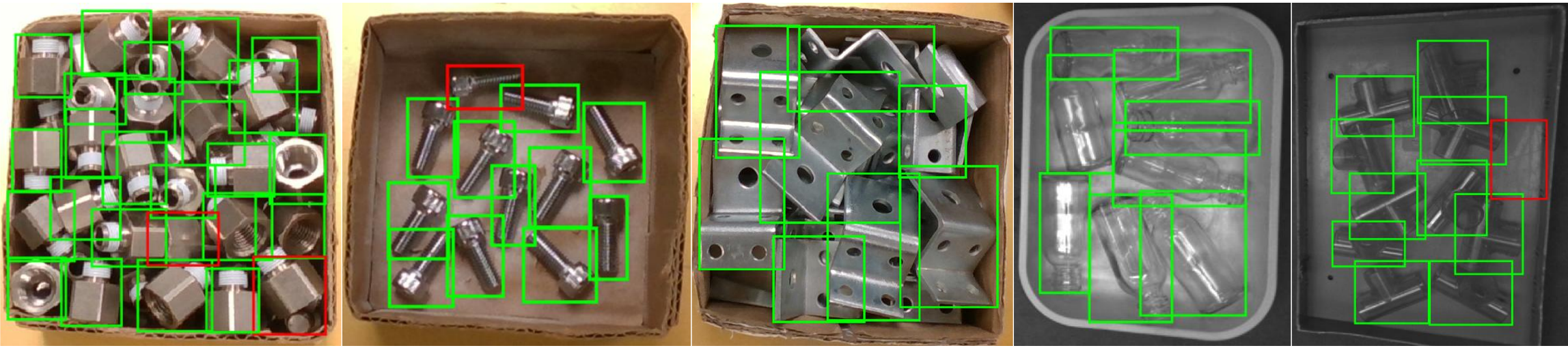}
    \caption{Qualitative results of YOLOv8 object detection on the ROBI and T-ROBI datasets. Detections with \textbf{IoU} $\boldsymbol{>} \mathbf{0.8}$ are considered correct (green), while others are false or inaccurate detections (red). From left to right: the object ``Tube Fitting'', ``Chrome Screw'', ``Zigzag'' from ROBI dataset and ``Bottle'', ``Pipe Fitting'' from T-ROBI dataset.}
    \label{fig:results_yolo}
  }
\end{figure*}

\begin{table*}[t]
\centering
\resizebox{0.7\textwidth}{!}{
\begin{tabular}{lcccc}
\toprule
\multirow{2}{*}{Dataset} & \multicolumn{2}{c}{IoU $>$ 0.7} & \multicolumn{2}{c}{IoU $>$ 0.8} \\ 
\cmidrule(lr){2-3} \cmidrule(lr){4-5}
 & Detection (\%) & False (\%) & Detection (\%) & False (\%) \\
\midrule
ROBI (Ensenso)     & 91.4 & 13.9 & 85.4 & 20.4 \\
ROBI (RealSense)   & 90.9 & 11.1 & 86.1 & 16.6 \\
T-ROBI             & 98.8 & 14.0 & 98.3 & 14.5 \\
\bottomrule
\end{tabular}}
\caption{YOLOv8 object detection performance on the ROBI and T-ROBI datasets under different IoU thresholds. The table reports the detection rate and false detection rate for both thresholds.}
\label{tab_yolo}
\end{table*}

\subsection{YOLO Detection Performance}
As described in Section~\ref{Sec:translation}, we first employ YOLOv8~\citep{sohan2024review} to detect bounding boxes of multiple identical objects. Since accurate detection forms the foundation for reliable pose estimation, we first evaluate the 2D detection performance of YOLOv8 before presenting the pose estimation evaluations. We train YOLOv8 exclusively on our synthetic dataset and evaluate it on the real-world ROBI and T-ROBI datasets, which contain multiple instances of identical objects. YOLO’s detection performance is assessed using the commonly adopted Intersection over Union (IoU) metric. Specifically, IoU thresholds of 0.7 and 0.8 are used for quantitative evaluation. Table~\ref{tab_yolo} reports the detection rate and false detection rate under both thresholds. “Detection (\%)” denotes the ratio of correctly detected objects to the total number of ground-truth instances, while “False (\%)” represents the proportion of false detections relative to the total number of detected objects.

As shown in Table~\ref{tab_yolo}, YOLOv8 achieves a high detection rate (above 90\%) and a low false rate (below 15\%) across all datasets when evaluated with the 0.7 IoU threshold. Even under the stricter 0.8 IoU criterion, the detection performance remains consistently strong. Representative results in Figure~\ref{fig:results_yolo} further confirm that YOLOv8 provides sufficiently accurate detections for the subsequent object pose estimation stages.

\subsection{Baselines and Implementations for Pose Estimation}
We quantitatively evaluate our approach against three prominent baselines: Multi-View 3D Keypoints (MV-3D-KP)~\citep{li2023multi} and two variants of CosyPose~\citep{labbe2020cosypose}. To ensure a fair comparison, all methods are trained only on the synthetic dataset (described in Section~\ref{sec_dataset}). During runtime, we utilize identical object bounding box detections and provide ground truth multi-view camera poses.

\begin{itemize}
    \item \textbf{MV-3D-KP}. Multi-View 3D Keypoints (MV-3D-KP)~\citep{li2023multi} builds upon the single-view approach of PVN3D~\citep{he2020pvn3d} and specializes in estimating 6D object poses by leveraging both RGB and depth data. MV-3D-KP provides excellent scalability, allowing for the incorporation of additional views that enhance accuracy and reduce uncertainty in pose estimation. As shown in~\citep{li2023multi}, this method demonstrates exceptional performance on the ROBI dataset, setting a high standard in the field.

    \item \textbf{CosyPose+PVNet}. CosyPose~\citep{labbe2020cosypose} is a multi-view pose fusion solution which takes the 6D object pose estimates from individual viewpoints as the input and optimizes the overall scene consistency. Note that, CosyPose is an offline batch-based solution that is agnostic to any particular pose estimator. In our implementation, we utilize a learning-based approach, Pixel-Wise Voting Network (PVNet)~\citep{peng2019pvnet}, to acquire the single view pose estimates. The PVNet approach first detects 2D keypoints and then solves a Perspective-n-Point (PnP) problem for pose estimation. This approach naturally deals with object occlusion and achieves remarkable performance.

    \item \textbf{CosyPose+LINE2D}. To provide single view pose estimates for CosyPose, we additionally utilize the LINE-2D pose estimator. In our implementation, we utilize the LINE-2D pose estimator with the same object center, object edge, and segmentation mask (from our MEC-Net). To feed the reliable single-view estimates to CosyPose, we use two strategies to obtain scale information. For the first strategy, we generate the templates at multiple distances during training (9 distances in our experiments) and perform standard template matching at inference time. This strategy can significantly improve the single view pose estimation performance by sacrificing run-time speed and is treated as the RGB version. For the second strategy, we directly use the depth images at inference time to acquire the object scale and refer to it as the RGB-D version.
\end{itemize}

We implement our MEC-Net using the PyTorch library, employing ResNet-18~\citep{he2016deep} as the backbone network. The MEC-Net is trained from scratch using the Adam optimizer~\citep{kingma2014adam}, with a batch size of 640 and a learning rate of 0.001 over 100 epochs on an RTX A6000 GPU. To ensure a fair comparison between MV-3D-KP and PVNet, we use the same ResNet-18 backbone and maintain consistent hyperparameters during training.
\\

\begin{figure*}[t]
\centering
\begin{subfigure}{\textwidth}
    \centering
    \includegraphics[width=0.9\linewidth]{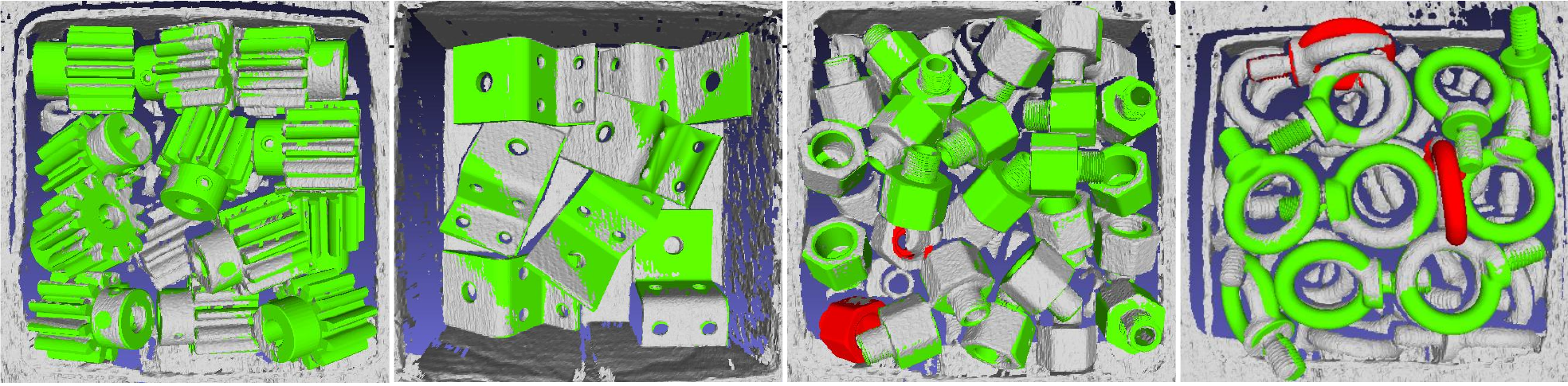}
    \caption{Object pose estimation results on \textbf{ROBI} dataset.}
    \label{fig_robi_results}
\end{subfigure}
\begin{subfigure}{\textwidth}
    \centering
    \includegraphics[width=0.9\linewidth]{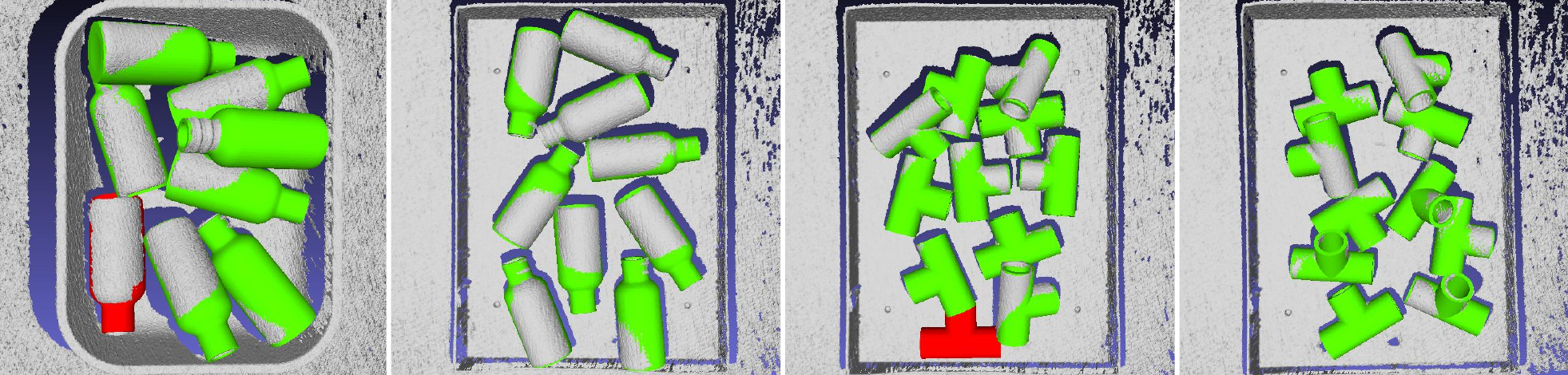}
    \caption{Object pose estimation results on \textbf{T-ROBI} dataset.}
    \label{fig_trobi_results}
\end{subfigure}
\caption{Qualitative results of our approach for the \textbf{ROBI}, \textbf{T-ROBI} datasets. Pose estimation performance is depicted using color coding: green indicates detections that satisfy the ADD* metric, while red indicates those that do not. The results are generated using 8 camera viewpoints. To enhance visualization, the estimated object poses are overlaid on the ground truth depth map.}
\label{qualitative_results_robi}
\end{figure*}

\subsection{Evaluation Metrics for Pose Estimation}
In our evaluation, we consider a ground truth pose only if its visibility score is larger than 75\%. We adopt two metrics to evaluate pose estimation performance: the symmetry-aware average model distance (\textbf{ADD*}) and the 5-mm/10-degree ($\mathbf{5\:mm, 10^{\circ}}$) metric. 

For objects with known geometric symmetries, the standard ADD metric~\citep{hinterstoisser2012model} may incorrectly penalize poses that are visually indistinguishable under the object’s symmetry group. A commonly used alternative, ADD-S, handles symmetry via nearest-neighbor matching but can tolerate large pose errors. To address this, we adopt a symmetry-aware variant of ADD, denoted as \textbf{ADD*}, that leverages the object’s known symmetry group. Let $\mathcal{S} = \{\mathbf{S}^1, \dots, \mathbf{S}^K\}$ denote the set of symmetry transformations that leave the object appearance unchanged. For each symmetry operation $\mathbf{S}^k \in \mathbb{SO}(3)$, we construct a symmetry-equivalent ground truth pose $\mathbf{R}_{\text{gt}}^k, \mathbf{t}_{\text{gt}}^k$ as:
\begin{equation}
\mathbf{R}_{\text{gt}}^k = \mathbf{R}_{\text{gt}} \mathbf{S}^{(k)} \:\:,\:\: \mathbf{t}_{\text{gt}}^k = \mathbf{t}_{\text{gt}}.
\end{equation}
Given the estimated orientation $\mathbf{R}$ and translation $\mathbf{t}$, the ADD error with respect to symmetry operation $\mathbf{S}^k$ is:
\begin{equation}
\mathrm{ADD}\left(k\right)
= \frac{1}{|\mathcal{M}|}
\sum_{x \in \mathcal{M}}
\left\|
(\mathbf{R}\mathbf{x} + \mathbf{t}) -
\left(\mathbf{R}_{\text{gt}}^k \mathbf{x} + \mathbf{t}_{\text{gt}}^k\right)
\right\|,
\end{equation}
where $\mathcal{M}$ is the set of 3D object model points. The final symmetry-aware error is defined as the minimum over all symmetry-adjusted ground truth poses:
\begin{equation}
\mathrm{ADD}^* = \min_{i \in \{1,\dots,K\}} \mathrm{ADD}\left(k\right).
\end{equation}
The ADD* metric enforces precise geometric correspondence while respecting object symmetry and reduces to standard ADD metric for non-symmetric objects. An object pose is considered correct if its ADD* is smaller than 10\% of the object diameter.

To further evaluate pose accuracy, we also use the stricter 5-mm/10-degree ($5\:mm$, $10^{\circ}$) metric. we reuse the symmetry-equivalent ground truth poses defined for ADD*. The rotation and translation errors are evaluated against each symmetry-adjusted pose ${\mathbf{R}_{\text{gt}}^k, \mathbf{t}_{\text{gt}}^k}$, and the estimate is considered correct if it satisfies both the 5-mm translation and 10-degree rotation thresholds for at least one pose.

\begin{table*}[t]
\resizebox{\textwidth}{!}{
\begin{threeparttable}
\begin{tabular}{|cc|ccccc|ccccc|}
\hline
\multicolumn{2}{|c|}{\multirow{3}{*}{Objects}}                                                            & \multicolumn{5}{c|}{\multirow{2}{*}{4 Views}}                                                                                                                                                                                                               & \multicolumn{5}{c|}{\multirow{2}{*}{8 Views}}                                                                                                                                                                                                               \\
\multicolumn{2}{|c|}{}                                                                                    & \multicolumn{5}{c|}{}                                                                                                                                                                                                                                       & \multicolumn{5}{c|}{}                                                                                                                                                                                                                                       \\ \cline{3-12} 
\multicolumn{2}{|c|}{}                                                                                    & \multicolumn{1}{c|}{\begin{tabular}[c]{@{}c@{}}CosyPose\\ +PVNet\end{tabular}} & \multicolumn{2}{c|}{\begin{tabular}[c]{@{}c@{}}CosyPose\\ +LINE2D\end{tabular}} & \multicolumn{1}{c|}{\begin{tabular}[c]{@{}c@{}}MV-\\ 3D-KP\end{tabular}} & Ours          & \multicolumn{1}{c|}{\begin{tabular}[c]{@{}c@{}}CosyPose\\ +PVNet\end{tabular}} & \multicolumn{2}{c|}{\begin{tabular}[c]{@{}c@{}}CosyPose\\ +LINE2D\end{tabular}} & \multicolumn{1}{c|}{\begin{tabular}[c]{@{}c@{}}MV-\\ 3D-KP\end{tabular}} & Ours          \\ \hline
\multicolumn{2}{|c|}{Input Modality}                                                                      & \multicolumn{1}{c|}{RGB}                                                       & \multicolumn{1}{c|}{RGB}               & \multicolumn{1}{c|}{RGBD}              & \multicolumn{1}{c|}{RGBD}                                                & RGB           & \multicolumn{1}{c|}{RGB}                                                       & \multicolumn{1}{c|}{RGB}               & \multicolumn{1}{c|}{RGBD}              & \multicolumn{1}{c|}{RGBD}                                                & RGB           \\ \hline
\multicolumn{1}{|c|}{\multirow{2}{*}{\begin{tabular}[c]{@{}c@{}}Tube\\ Fitting\end{tabular}}}    & ADD*    & \multicolumn{1}{c|}{39.4}                                                      & \multicolumn{1}{c|}{32.5}              & \multicolumn{1}{c|}{74.8}              & \multicolumn{1}{c|}{\textbf{94.0}}                                       & \underline{89.4}          & \multicolumn{1}{c|}{61.6}                                                      & \multicolumn{1}{c|}{50.3}              & \multicolumn{1}{c|}{91.4}              & \multicolumn{1}{c|}{\textbf{96.0}}                                       & \underline{94.0}          \\ \cline{2-12} 
\multicolumn{1}{|c|}{}                                                                           & (5,10) & \multicolumn{1}{c|}{47.7}                                                      & \multicolumn{1}{c|}{45.7}              & \multicolumn{1}{c|}{76.2}              & \multicolumn{1}{c|}{\textbf{95.4}}                                       & \underline{88.1}          & \multicolumn{1}{c|}{64.9}                                                      & \multicolumn{1}{c|}{71.5}              & \multicolumn{1}{c|}{\underline{94.7}}              & \multicolumn{1}{c|}{\textbf{96.0}}                                       & 92.0          \\ \hline
\multicolumn{1}{|c|}{\multirow{2}{*}{\begin{tabular}[c]{@{}c@{}}Chrome\\ Screw\end{tabular}}}    & ADD*    & \multicolumn{1}{c|}{17.4}                                                      & \multicolumn{1}{c|}{55.7}              & \multicolumn{1}{c|}{73.0}              & \multicolumn{1}{c|}{\textbf{90.8}}                                       & \underline{86.7}          & \multicolumn{1}{c|}{24.4}                                                      & \multicolumn{1}{c|}{70.1}              & \multicolumn{1}{c|}{88.5}              & \multicolumn{1}{c|}{\underline{91.9}}                                                & \textbf{93.7} \\ \cline{2-12} 
\multicolumn{1}{|c|}{}                                                                           & (5,10) & \multicolumn{1}{c|}{18.6}                                                      & \multicolumn{1}{c|}{63.2}              & \multicolumn{1}{c|}{78.2}              & \multicolumn{1}{c|}{\textbf{88.5}}                                       & \underline{85.1}          & \multicolumn{1}{c|}{30.8}                                                      & \multicolumn{1}{c|}{78.7}              & \multicolumn{1}{c|}{90.2}              & \multicolumn{1}{c|}{\textbf{90.8}}                                       & 90.2          \\ \hline
\multicolumn{1}{|c|}{\multirow{2}{*}{\begin{tabular}[c]{@{}c@{}}Eye\\ Bolt\end{tabular}}}        & ADD*    & \multicolumn{1}{c|}{21.6}                                                      & \multicolumn{1}{c|}{35.1}              & \multicolumn{1}{c|}{85.1}              & \multicolumn{1}{c|}{\textbf{93.2}}                                       & \textbf{93.2}          & \multicolumn{1}{c|}{46.0}                                                      & \multicolumn{1}{c|}{79.7}              & \multicolumn{1}{c|}{93.2}              & \multicolumn{1}{c|}{\textbf{94.6}}                                       & \textbf{94.6}          \\ \cline{2-12} 
\multicolumn{1}{|c|}{}                                                                           & (5,10) & \multicolumn{1}{c|}{12.2}                                                      & \multicolumn{1}{c|}{27.0}              & \multicolumn{1}{c|}{\underline{78.4}}              & \multicolumn{1}{c|}{\textbf{87.8}}                                       & 67.6          & \multicolumn{1}{c|}{31.1}                                                      & \multicolumn{1}{c|}{64.9}              & \multicolumn{1}{c|}{\underline{83.8}}              & \multicolumn{1}{c|}{\textbf{85.1}}                                       & 75.8          \\ \hline
\multicolumn{1}{|c|}{\multirow{2}{*}{Gear\tnote{*}}}                                                      & ADD*    & \multicolumn{1}{c|}{50.6}                                                      & \multicolumn{1}{c|}{25.9}              & \multicolumn{1}{c|}{80.2}              & \multicolumn{1}{c|}{\underline{85.2}}                                                & \textbf{91.4} & \multicolumn{1}{c|}{71.6}                                                      & \multicolumn{1}{c|}{43.2}              & \multicolumn{1}{c|}{88.9}              & \multicolumn{1}{c|}{\underline{93.8}}                                                & \textbf{97.5} \\ \cline{2-12} 
\multicolumn{1}{|c|}{}                                                                           & (5,10) & \multicolumn{1}{c|}{34.6}                                                      & \multicolumn{1}{c|}{29.6}              & \multicolumn{1}{c|}{79.0}              & \multicolumn{1}{c|}{\textbf{85.2}}                                                & \textbf{85.2} & \multicolumn{1}{c|}{49.4}                                                      & \multicolumn{1}{c|}{45.7}              & \multicolumn{1}{c|}{\underline{92.6}}              & \multicolumn{1}{c|}{91.4}                                                & \textbf{93.8} \\ \hline
\multicolumn{1}{|c|}{\multirow{2}{*}{Zigzag}}                                                    & ADD*    & \multicolumn{1}{c|}{89.7}                                                      & \multicolumn{1}{c|}{65.5}              & \multicolumn{1}{c|}{87.9}              & \multicolumn{1}{c|}{\textbf{96.6}}                                       & \underline{94.8}          & \multicolumn{1}{c|}{89.7}                                                      & \multicolumn{1}{c|}{77.6}              & \multicolumn{1}{c|}{96.6}              & \multicolumn{1}{c|}{96.6}                                                & \textbf{98.3} \\ \cline{2-12} 
\multicolumn{1}{|c|}{}                                                                           & (5,10) & \multicolumn{1}{c|}{82.8}                                                      & \multicolumn{1}{c|}{37.9}              & \multicolumn{1}{c|}{75.9}              & \multicolumn{1}{c|}{\textbf{93.1}}                                       & \underline{89.7}          & \multicolumn{1}{c|}{86.2}                                                      & \multicolumn{1}{c|}{63.8}              & \multicolumn{1}{c|}{93.1}              & \multicolumn{1}{c|}{\textbf{96.6}}                                       & 93.1          \\ \hline
\multicolumn{1}{|c|}{\multirow{2}{*}{\begin{tabular}[c]{@{}c@{}}DIN\\ Connector\end{tabular}}}   & ADD*    & \multicolumn{1}{c|}{13.3}                                                      & \multicolumn{1}{c|}{15.6}              & \multicolumn{1}{c|}{57.8}              & \multicolumn{1}{c|}{\textbf{90.6}}                                       & \underline{69.5}          & \multicolumn{1}{c|}{28.1}                                                      & \multicolumn{1}{c|}{24.2}              & \multicolumn{1}{c|}{64.1}              & \multicolumn{1}{c|}{\textbf{93.8}}                                       & \underline{73.4}          \\ \cline{2-12} 
\multicolumn{1}{|c|}{}                                                                           & (5,10) & \multicolumn{1}{c|}{18.8}                                                      & \multicolumn{1}{c|}{12.5}              & \multicolumn{1}{c|}{46.1}              & \multicolumn{1}{c|}{\textbf{84.4}}                                       & \underline{53.9}          & \multicolumn{1}{c|}{32.0}                                                      & \multicolumn{1}{c|}{23.4}              & \multicolumn{1}{c|}{51.6}              & \multicolumn{1}{c|}{\textbf{93.0}}                                       & \underline{59.4}          \\ \hline
\multicolumn{1}{|c|}{\multirow{2}{*}{\begin{tabular}[c]{@{}c@{}}D-Sub\\ Connector\tnote{\textdagger}\end{tabular}}} & ADD*    & \multicolumn{1}{c|}{11.2}                                                      & \multicolumn{1}{c|}{9.9}               & \multicolumn{1}{c|}{55.3}              & \multicolumn{1}{c|}{\textbf{92.5}}                                       & \underline{79.5}          & \multicolumn{1}{c|}{18.0}                                                      & \multicolumn{1}{c|}{15.5}              & \multicolumn{1}{c|}{63.3}              & \multicolumn{1}{c|}{\textbf{95.7}}                                       & \underline{84.5}          \\ \cline{2-12} 
\multicolumn{1}{|c|}{}                                                                           & (5,10) & \multicolumn{1}{c|}{11.2}                                                      & \multicolumn{1}{c|}{11.2}              & \multicolumn{1}{c|}{39.1}              & \multicolumn{1}{c|}{\textbf{83.2}}                                       & \underline{47.2}          & \multicolumn{1}{c|}{16.8}                                                      & \multicolumn{1}{c|}{11.2}              & \multicolumn{1}{c|}{41.6}              & \multicolumn{1}{c|}{\textbf{91.3}}                                       & \underline{55.9}          \\ \hline
\multicolumn{1}{|c|}{\multirow{2}{*}{ALL}}                                                       & ADD*    & \multicolumn{1}{c|}{34.7}                                                      & \multicolumn{1}{c|}{34.3}              & \multicolumn{1}{c|}{73.4}              & \multicolumn{1}{c|}{\textbf{91.8}}                                       & \underline{86.4}          & \multicolumn{1}{c|}{48.5}                                                      & \multicolumn{1}{c|}{51.5}              & \multicolumn{1}{c|}{83.7}              & \multicolumn{1}{c|}{\textbf{94.6}}                                       & \underline{90.9}          \\ \cline{2-12} 
\multicolumn{1}{|c|}{}                                                                           & (5,10) & \multicolumn{1}{c|}{32.3}                                                      & \multicolumn{1}{c|}{32.4}              & \multicolumn{1}{c|}{67.6}              & \multicolumn{1}{c|}{\textbf{88.2}}                                       & \underline{73.8}          & \multicolumn{1}{c|}{44.5}                                                      & \multicolumn{1}{c|}{51.3}              & \multicolumn{1}{c|}{78.2}              & \multicolumn{1}{c|}{\textbf{92.0}}                                       & \underline{80.0}          \\ \hline
\end{tabular}
 \end{threeparttable}}
    \begin{tablenotes}
      \item *In our evaluation, we treat the object ``Gear'' as symmetric about the Z-axis with an order of 12.
      \item \textdagger In our evaluation, we treat the object ``D-Sub Connector'' as symmetric about the Z-axis with an order of 2. 
    \end{tablenotes}
\caption[Detection rates of RGB-based object pose estimation on Ensenso test set from ROBI dataset]{Detection rates of 6D object pose estimation on \textbf{Ensenso} test set from \textbf{ROBI} dataset, evaluated with the metrics of ADD* and ($5\:mm$, $10^{\circ}$). There are a total of nine scenes for each object.}
\label{tab_ensenso_results}
\end{table*}

\begin{table*}[t]
\resizebox{\textwidth}{!}{
 \begin{threeparttable}
\begin{tabular}{|cc|ccccc|ccccc|}
\hline
\multicolumn{2}{|c|}{\multirow{3}{*}{Objects}}                                                            & \multicolumn{5}{c|}{\multirow{2}{*}{4 Views}}                                                                                                                                                                                                               & \multicolumn{5}{c|}{\multirow{2}{*}{8 Views}}                                                                                                                                                                                                               \\
\multicolumn{2}{|c|}{}                                                                                    & \multicolumn{5}{c|}{}                                                                                                                                                                                                                                       & \multicolumn{5}{c|}{}                                                                                                                                                                                                                                       \\ \cline{3-12} 
\multicolumn{2}{|c|}{}                                                                                    & \multicolumn{1}{c|}{\begin{tabular}[c]{@{}c@{}}CosyPose\\ +PVNet\end{tabular}} & \multicolumn{2}{c|}{\begin{tabular}[c]{@{}c@{}}CosyPose\\ +LINE2D\end{tabular}} & \multicolumn{1}{c|}{\begin{tabular}[c]{@{}c@{}}MV-\\ 3D-KP\end{tabular}} & Ours          & \multicolumn{1}{c|}{\begin{tabular}[c]{@{}c@{}}CosyPose\\ +PVNet\end{tabular}} & \multicolumn{2}{c|}{\begin{tabular}[c]{@{}c@{}}CosyPose\\ +LINE2D\end{tabular}} & \multicolumn{1}{c|}{\begin{tabular}[c]{@{}c@{}}MV-\\ 3D-KP\end{tabular}} & Ours          \\ \hline
\multicolumn{2}{|c|}{Input Modality}                                                                      & \multicolumn{1}{c|}{RGB}                                                       & \multicolumn{1}{c|}{RGB}           & \multicolumn{1}{c|}{RGBD}                  & \multicolumn{1}{c|}{RGBD}                                                & RGB           & \multicolumn{1}{c|}{RGB}                                                       & \multicolumn{1}{c|}{RGB}           & \multicolumn{1}{c|}{RGBD}                  & \multicolumn{1}{c|}{RGBD}                                                & RGB           \\ \hline
\multicolumn{1}{|c|}{\multirow{2}{*}{\begin{tabular}[c]{@{}c@{}}Tube\\ Fitting\end{tabular}}}    & ADD*    & \multicolumn{1}{c|}{26.7}                                                      & \multicolumn{1}{c|}{27.9}          & \multicolumn{1}{c|}{70.6}                  & \multicolumn{1}{c|}{\underline{79.4}}                                                & \textbf{86.8} & \multicolumn{1}{c|}{47.1}                                                      & \multicolumn{1}{c|}{69.1}          & \multicolumn{1}{c|}{83.9}                  & \multicolumn{1}{c|}{\textbf{86.8}}                                                & \underline{85.3} \\ \cline{2-12} 
\multicolumn{1}{|c|}{}                                                                           & (5,10) & \multicolumn{1}{c|}{36.8}                                                      & \multicolumn{1}{c|}{48.5}          & \multicolumn{1}{c|}{72.1}                  & \multicolumn{1}{c|}{\underline{76.5}}                                                & \textbf{79.4} & \multicolumn{1}{c|}{44.2}                                                      & \multicolumn{1}{c|}{82.3}          & \multicolumn{1}{c|}{\underline{85.3}}                  & \multicolumn{1}{c|}{83.8}                                                & \textbf{91.2} \\ \hline
\multicolumn{1}{|c|}{\multirow{2}{*}{\begin{tabular}[c]{@{}c@{}}Chrome\\ Screw\end{tabular}}}    & ADD*    & \multicolumn{1}{c|}{10.0}                                                      & \multicolumn{1}{c|}{58.6}          & \multicolumn{1}{c|}{68.5}                  & \multicolumn{1}{c|}{\underline{87.1}}                                                & \textbf{92.9} & \multicolumn{1}{c|}{30.0}                                                      & \multicolumn{1}{c|}{77.1}          & \multicolumn{1}{c|}{80.0}                  & \multicolumn{1}{c|}{\textbf{92.9}}                                                & \textbf{92.9} \\ \cline{2-12} 
\multicolumn{1}{|c|}{}                                                                           & (5,10) & \multicolumn{1}{c|}{10.0}                                                      & \multicolumn{1}{c|}{64.3}          & \multicolumn{1}{c|}{\underline{82.9}}         & \multicolumn{1}{c|}{\textbf{84.3}}                                                & 77.1          & \multicolumn{1}{c|}{42.9}                                                      & \multicolumn{1}{c|}{85.7}          & \multicolumn{1}{c|}{\textbf{94.3}}         & \multicolumn{1}{c|}{\underline{91.4}}                                                & 87.1          \\ \hline
\multicolumn{1}{|c|}{\multirow{2}{*}{\begin{tabular}[c]{@{}c@{}}Eye\\ Bolt\end{tabular}}}        & ADD*    & \multicolumn{1}{c|}{17.7}                                                      & \multicolumn{1}{c|}{58.8}          & \multicolumn{1}{c|}{76.5}                  & \multicolumn{1}{c|}{\underline{85.3}}                                                & \textbf{94.1} & \multicolumn{1}{c|}{38.2}                                                      & \multicolumn{1}{c|}{73.5}          & \multicolumn{1}{c|}{\textbf{94.1}}         & \multicolumn{1}{c|}{85.3}                                                & \textbf{94.1} \\ \cline{2-12} 
\multicolumn{1}{|c|}{}                                                                           & (5,10) & \multicolumn{1}{c|}{17.7}                                                      & \multicolumn{1}{c|}{41.2}          & \multicolumn{1}{c|}{\underline{67.6}}                  & \multicolumn{1}{c|}{\textbf{82.3}}                                       & 55.9          & \multicolumn{1}{c|}{29.4}                                                      & \multicolumn{1}{c|}{61.8}          & \multicolumn{1}{c|}{\textbf{91.2}}         & \multicolumn{1}{c|}{\underline{85.3}}                                                & 76.5          \\ \hline
\multicolumn{1}{|c|}{\multirow{2}{*}{Gear\tnote{*}}}                                                      & ADD*    & \multicolumn{1}{c|}{38.9}                                                      & \multicolumn{1}{c|}{36.1}          & \multicolumn{1}{c|}{83.3}                  & \multicolumn{1}{c|}{\underline{88.8}}                                                & \textbf{94.4} & \multicolumn{1}{c|}{44.4}                                                      & \multicolumn{1}{c|}{55.6}          & \multicolumn{1}{c|}{\textbf{97.2}}         & \multicolumn{1}{c|}{91.7}                                                & \textbf{97.2} \\ \cline{2-12} 
\multicolumn{1}{|c|}{}                                                                           & (5,10) & \multicolumn{1}{c|}{27.8}                                                      & \multicolumn{1}{c|}{38.9}          & \multicolumn{1}{c|}{\underline{77.8}}                  & \multicolumn{1}{c|}{61.1}                                                & \textbf{86.1} & \multicolumn{1}{c|}{30.6}                                                      & \multicolumn{1}{c|}{58.3}          & \multicolumn{1}{c|}{\textbf{94.4}}         & \multicolumn{1}{c|}{63.9}                                                & \underline{88.9}          \\ \hline
\multicolumn{1}{|c|}{\multirow{2}{*}{Zigzag}}                                                    & ADD*    & \multicolumn{1}{c|}{60.7}                                                      & \multicolumn{1}{c|}{42.9}          & \multicolumn{1}{c|}{78.6}                  & \multicolumn{1}{c|}{\textbf{96.4}}                                       & 89.3          & \multicolumn{1}{c|}{85.7}                                                      & \multicolumn{1}{c|}{71.4}          & \multicolumn{1}{c|}{92.9}                  & \multicolumn{1}{c|}{92.9}                                       & \textbf{96.4} \\ \cline{2-12} 
\multicolumn{1}{|c|}{}                                                                           & (5,10) & \multicolumn{1}{c|}{53.6}                                                      & \multicolumn{1}{c|}{21.4}          & \multicolumn{1}{c|}{71.4}                  & \multicolumn{1}{c|}{\textbf{96.4}}                                       & \underline{85.7}          & \multicolumn{1}{c|}{82.1}                                                      & \multicolumn{1}{c|}{64.3}          & \multicolumn{1}{c|}{92.9}                  & \multicolumn{1}{c|}{92.9}                                       & 92.9          \\ \hline
\multicolumn{1}{|c|}{\multirow{2}{*}{\begin{tabular}[c]{@{}c@{}}DIN\\ Connector\end{tabular}}}   & ADD*    & \multicolumn{1}{c|}{11.5}                                                      & \multicolumn{1}{c|}{3.8}           & \multicolumn{1}{c|}{36.5}                  & \multicolumn{1}{c|}{\textbf{82.7}}                                       & \underline{51.9}          & \multicolumn{1}{c|}{15.4}                                                      & \multicolumn{1}{c|}{15.4}          & \multicolumn{1}{c|}{51.9}                  & \multicolumn{1}{c|}{\textbf{84.6}}                                       & \underline{82.7}          \\ \cline{2-12} 
\multicolumn{1}{|c|}{}                                                                           & (5,10) & \multicolumn{1}{c|}{13.5}                                                      & \multicolumn{1}{c|}{1.9}           & \multicolumn{1}{c|}{30.8}                  & \multicolumn{1}{c|}{\textbf{75.0}}                                       & \underline{32.7}          & \multicolumn{1}{c|}{26.9}                                                      & \multicolumn{1}{c|}{9.6}           & \multicolumn{1}{c|}{34.6}                  & \multicolumn{1}{c|}{\textbf{84.6}}                                       & \underline{57.7}          \\ \hline
\multicolumn{1}{|c|}{\multirow{2}{*}{\begin{tabular}[c]{@{}c@{}}D-Sub\\ Connector\tnote{\textdagger}\end{tabular}}} & ADD*    & \multicolumn{1}{c|}{8.3}                                                       & \multicolumn{1}{c|}{6.9}           & \multicolumn{1}{c|}{40.3}                  & \multicolumn{1}{c|}{\textbf{88.9}}                                       & \underline{70.8}          & \multicolumn{1}{c|}{20.8}                                                      & \multicolumn{1}{c|}{9.7}           & \multicolumn{1}{c|}{45.8}                  & \multicolumn{1}{c|}{\textbf{90.3}}                                                & \underline{81.9} \\ \cline{2-12} 
\multicolumn{1}{|c|}{}                                                                           & (5,10) & \multicolumn{1}{c|}{9.7}                                                       & \multicolumn{1}{c|}{6.9}           & \multicolumn{1}{c|}{18.1}                  & \multicolumn{1}{c|}{\textbf{44.4}}                                       & \underline{31.9}          & \multicolumn{1}{c|}{18.1}                                                      & \multicolumn{1}{c|}{8.3}           & \multicolumn{1}{c|}{33.3}                  & \multicolumn{1}{c|}{\textbf{55.6}}                                                & \underline{43.1} \\ \hline
\multicolumn{1}{|c|}{\multirow{2}{*}{ALL}}                                                       & ADD*    & \multicolumn{1}{c|}{24.8}                                                      & \multicolumn{1}{c|}{33.6}          & \multicolumn{1}{c|}{64.9}                  & \multicolumn{1}{c|}{\textbf{86.7}}                                       & \underline{82.9}          & \multicolumn{1}{c|}{40.2}                                                      & \multicolumn{1}{c|}{53.1}          & \multicolumn{1}{c|}{78.0}                  & \multicolumn{1}{c|}{\underline{89.2}}                                                & \textbf{90.1} \\ \cline{2-12} 
\multicolumn{1}{|c|}{}                                                                           & (5,10) & \multicolumn{1}{c|}{24.2}                                                      & \multicolumn{1}{c|}{31.9}          & \multicolumn{1}{c|}{60.1}                  & \multicolumn{1}{c|}{\textbf{74.3}}                                       & \underline{64.1}          & \multicolumn{1}{c|}{39.2}                                                      & \multicolumn{1}{c|}{52.9}          & \multicolumn{1}{c|}{75.1}                  & \multicolumn{1}{c|}{\textbf{79.6}}                                                & \underline{76.8} \\ \hline
\end{tabular}
 \end{threeparttable}}
\caption[Detection rates of RGB-based object pose estimation on RealSense test set from ROBI dataset]{Detection rates of 6D object pose estimation on \textbf{RealSense} test set from \textbf{ROBI} dataset, evaluated with the metrics of ADD* and ($5\:mm$, $10^{\circ}$). There are a total of four scenes for each object.}
\label{tab_realsense_results}
\end{table*}

\subsection{Pose Estimation Results on ROBI}
We conduct experiments on the ROBI dataset with a variable number of viewpoints (4 and 8), with the viewpoints carefully chosen to provide broad coverage of the scene. Figure~\ref{fig_robi_results} illustrates the qualitative superiority of our approach. Quantitative results on the Ensenso and RealSense test sets are presented in Tables~\ref{tab_ensenso_results} and~\ref{tab_realsense_results}, respectively. The results show our method outperforms the RGB baseline by a wide margin, and is competitive with the RGB-D approaches, without the need for depth measurements.

In the Ensenso test set, it is noteworthy that ``MV-3D-KP'' demonstrates exceptional performance, achieving state-of-the-art results on the ROBI dataset. This success is largely attributed to the high-quality depth maps produced by the Ensenso 3D camera. Specifically, when utilizing RGB-D data, the ``MV-3D-KP'' method achieves an overall detection rate of $91.8\%$ using four views and $94.6\%$ using eight views, as measured by the ADD* metric. Additionally, it achieves an overall detection rate of $88.2\%$ with four views and $92.0\%$ with eight views using the ($5\:mm$, $10^{\circ}$) metric. In comparison, despite relying solely on RGB data, our approach demonstrates competitive performance, with detection rates only $5.4\%$ and $3.7\%$ lower than MV-3D-KP for four-view and eight-view data, respectively, as measured by the ADD* metric. When utilizing only RGB images, our approach significantly outperforms both ``CosyPose+PVNet'' and ``CosyPose+LINE2D'', achieving margins of at least 51.7\% and 39.4\% for the 4-view and 8-view configurations, respectively, as measured by the ADD* metric. With the availability of depth data, the performance of ``CosyPose+LINE2D'' shows substantial improvement, representing its upper bound. In contrast, our method exceeds this upper bound by a clear margin, achieving detection rates that are 13.0\% and 7.2\% higher on the 4-view and 8-view test sets, respectively, with the ADD* metric. A similar margin is observed when using the ($5\:mm$, $10^{\circ}$) metric.

\begin{table*}[t]
\resizebox{\textwidth}{!}{
\begin{threeparttable}
\begin{tabular}{|cc|ccccc|ccccc|}
\hline
\multicolumn{2}{|c|}{\multirow{3}{*}{Objects}}                                                         & \multicolumn{5}{c|}{\multirow{2}{*}{4 Views}}                                                                                                                                                                                                               & \multicolumn{5}{c|}{\multirow{2}{*}{8 Views}}                                                                                                                                                                                                               \\
\multicolumn{2}{|c|}{}                                                                                 & \multicolumn{5}{c|}{}                                                                                                                                                                                                                                       & \multicolumn{5}{c|}{}                                                                                                                                                                                                                                       \\ \cline{3-12} 
\multicolumn{2}{|c|}{}                                                                                 & \multicolumn{1}{c|}{\begin{tabular}[c]{@{}c@{}}CosyPose\\ +PVNet\end{tabular}} & \multicolumn{2}{c|}{\begin{tabular}[c]{@{}c@{}}CosyPose\\ +LINE2D\end{tabular}} & \multicolumn{1}{c|}{\begin{tabular}[c]{@{}c@{}}MV-\\ 3D-KP\end{tabular}} & Ours          & \multicolumn{1}{c|}{\begin{tabular}[c]{@{}c@{}}CosyPose\\ +PVNet\end{tabular}} & \multicolumn{2}{c|}{\begin{tabular}[c]{@{}c@{}}CosyPose\\ +LINE2D\end{tabular}} & \multicolumn{1}{c|}{\begin{tabular}[c]{@{}c@{}}MV-\\ 3D-KP\end{tabular}} & Ours          \\ \hline
\multicolumn{2}{|c|}{Input Modality}                                                                   & \multicolumn{1}{c|}{RGB}                                                       & \multicolumn{1}{c|}{RGB}               & \multicolumn{1}{c|}{RGBD}              & \multicolumn{1}{c|}{RGBD}                                                & RGB           & \multicolumn{1}{c|}{RGB}                                                       & \multicolumn{1}{c|}{RGB}               & \multicolumn{1}{c|}{RGBD}              & \multicolumn{1}{c|}{RGBD}                                                & RGB           \\ \hline
\multicolumn{1}{|c|}{\multirow{2}{*}{Bottle}}                                                 & ADD*    & \multicolumn{1}{c|}{32.7}                                                      & \multicolumn{1}{c|}{\underline{38.5}}              & \multicolumn{1}{c|}{1.9}               & \multicolumn{1}{c|}{3.8}                                                 & \textbf{90.4} & \multicolumn{1}{c|}{48.1}                                                      & \multicolumn{1}{c|}{\underline{53.8}}              & \multicolumn{1}{c|}{3.8}               & \multicolumn{1}{c|}{1.9}                                                 & \textbf{90.4} \\ \cline{2-12} 
\multicolumn{1}{|c|}{}                                                                        & (5,10) & \multicolumn{1}{c|}{\underline{17.3}}                                                      & \multicolumn{1}{c|}{13.5}              & \multicolumn{1}{c|}{1.9}               & \multicolumn{1}{c|}{3.8}                                                 & \textbf{73.1} & \multicolumn{1}{c|}{28.9}                                                      & \multicolumn{1}{c|}{\underline{38.5}}              & \multicolumn{1}{c|}{3.8}               & \multicolumn{1}{c|}{1.9}                                                 & \textbf{75.0} \\ \hline
\multicolumn{1}{|c|}{\multirow{2}{*}{\begin{tabular}[c]{@{}c@{}}Pipe\\ Fitting\end{tabular}}} & ADD*    & \multicolumn{1}{c|}{30.4}                                                      & \multicolumn{1}{c|}{51.8}              & \multicolumn{1}{c|}{28.6}              & \multicolumn{1}{c|}{\underline{67.9}}                                                & \textbf{96.4} & \multicolumn{1}{c|}{39.3}                                                      & \multicolumn{1}{c|}{\underline{71.4}}              & \multicolumn{1}{c|}{37.5}              & \multicolumn{1}{c|}{64.3}                                                & \textbf{100.0}  \\ \cline{2-12} 
\multicolumn{1}{|c|}{}                                                                        & (5,10) & \multicolumn{1}{c|}{14.3}                                                      & \multicolumn{1}{c|}{35.7}              & \multicolumn{1}{c|}{12.5}              & \multicolumn{1}{c|}{\underline{60.7}}                                                & \textbf{85.7} & \multicolumn{1}{c|}{28.6}                                                      & \multicolumn{1}{c|}{41.7}              & \multicolumn{1}{c|}{17.9}              & \multicolumn{1}{c|}{\underline{62.5}}                                                & \textbf{87.5} \\ \hline
\multicolumn{1}{|c|}{\multirow{2}{*}{ALL}}                                                    & ADD*    & \multicolumn{1}{c|}{31.0}                                                      & \multicolumn{1}{c|}{\underline{45.2}}              & \multicolumn{1}{c|}{15.3}              & \multicolumn{1}{c|}{35.9}                                                & \textbf{93.4} & \multicolumn{1}{c|}{45.6}                                                      & \multicolumn{1}{c|}{\underline{62.6}}              & \multicolumn{1}{c|}{20.7}              & \multicolumn{1}{c|}{33.1}                                                & \textbf{95.2} \\ \cline{2-12} 
\multicolumn{1}{|c|}{}                                                                        & (5,10) & \multicolumn{1}{c|}{15.8}                                                      & \multicolumn{1}{c|}{24.6}              & \multicolumn{1}{c|}{7.2}               & \multicolumn{1}{c|}{\underline{32.3}}                                                & \textbf{79.4} & \multicolumn{1}{c|}{26.8}                                                      & \multicolumn{1}{c|}{\underline{40.1}}              & \multicolumn{1}{c|}{10.9}              & \multicolumn{1}{c|}{32.2}                                                & \textbf{81.3} \\ \hline
\end{tabular}
 \end{threeparttable}}
\caption{Detection rates of 6D object pose estimation on \textbf{T-ROBI} dataset, evaluated with the metrics of ADD* and ($5\:mm$, $10^{\circ}$). There are a total of six scenes for each object.}
\label{tab_t_robi_results}
\end{table*}

\begin{table*}[b]
\centering
\resizebox{0.82\textwidth}{!}{
\begin{tabular}{llccccccc}
\toprule
\multirow{3}{*}{Metric} & \multirow{3}{*}{\:\:\:\:\:\:\:\:\:\:\:\:Method} 
    & \multicolumn{7}{c}{Objects} \\ 
\cmidrule(lr){3-9}
& & Ball & $\text{Bottle}_0$ & $\text{Cup}_0$ & $\text{Mug}_4$ & Heart & Tree & ALL \\
\midrule
\multirow{2}{*}{ADD*} 
 & KeyPose+CosyPose & 90.6 & \textbf{100} & 96.9 & 96.9 & 53.1 & 81.3 & 86.5 \\
 & \:\:\:\:\:\:\:\:\:\:\:\:\:\:Ours             & \textbf{100} & \textbf{100} & \textbf{100} & \textbf{100} & \textbf{87.5} & \textbf{100} & \textbf{97.9} \\
\midrule
\multirow{2}{*}{(5,10)} 
 & KeyPose+CosyPose & 90.6 & \textbf{87.5} & 62.5 & 28.1 & 34.4 & 34.4 & 56.3 \\
 & \:\:\:\:\:\:\:\:\:\:\:\:\:\:Ours             & \textbf{96.9} & \textbf{87.5} & \textbf{100} & \textbf{84.4} & \textbf{71.9} & \textbf{90.6} & \textbf{88.6} \\
\bottomrule
\end{tabular}}
\caption{
Comparison of our method with KeyPose+CosyPose on the \textbf{TOD} dataset~\citep{liu2020keypose}. Evaluations use four stereo pairs (eight viewpoints per scene) with ADD* and ($5\:\text{mm}, 10^{\circ}$) metrics. \textbf{Bold} marks the best result in each column.}
\label{tab_keypose}
\end{table*}

In the RealSense test set, the degraded quality of depth data presents challenges for both the ``MV-3D-KP'' and ``CosyPose+LINE2D'' (RGB-D version) methods. In contrast and as expected, our approach maintains a comparable detection rate. Specifically, for the 4-view configuration, our approach exhibits only a slight decrease in performance compared to the ``MV-3D-KP'' by $3.8\%$ and $10.2\%$ using the ADD* and the ($5\:mm$, $10^{\circ}$) metric. With the 8-view configuration, our approach achieves the best performance of $90.1\%$ using the ADD* metric and is only $2.8\%$ lower than ``MV-3D-KP'' under the ($5\:mm$, $10^{\circ}$) metric.

\subsection{Pose Estimation Results on T-ROBI}
Table~\ref{tab_t_robi_results} presents the object pose estimation results on our T-ROBI dataset, where our approach demonstrates clear superiority. It significantly outperforms ``CosyPose'' (all variants) and ``MV-3D-KP'' by a substantial margin. Using the ADD* metric, our method demonstrates an impressive overall detection rate of 93.4\% for the 4-view configuration and 95.2\% for the 8-view configuration. When evaluated with the ($5\:mm$, $10^{\circ}$) metric, it achieves detection rates of 79.4\% with 4 views and 81.3\% with 8 views, highlighting its robustness in handling transparent objects. Figure~\ref{fig_trobi_results} further illustrates the strong performance of our approach on the T-ROBI dataset, with accurate pose estimation.

In contrast, the ``MV-3D-KP'' and ``CosyPose+LINE2D'' (RGB-D version) approaches show low detection rates, largely due to significant depth missing and inaccuracies. These results highlight the advantage of our RGB-only approach for transparent objects that typically challenge depth-based methods.

\begin{figure*}[t]
\centering
  \parbox{0.97\linewidth}{%
    \centering
    \includegraphics[width=\linewidth]{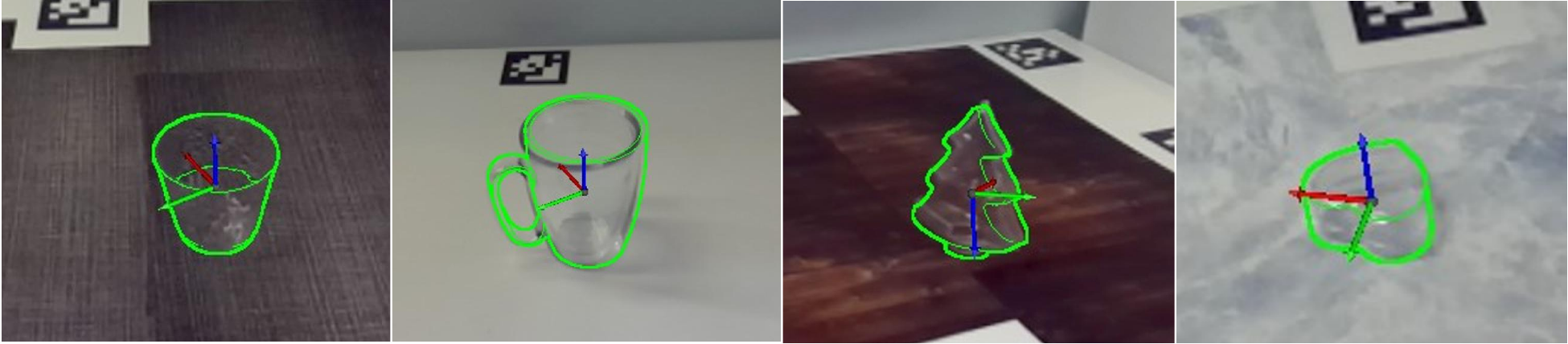}
    \caption{Qualitative results of our approach on the testing and validation sets of the \textbf{TOD}. The results are generated using 4 stereo pairs (8 viewpoints). From left to right: $\text{Mug}_4$, $\text{Cup}_0$, Tree, and Heart.}
    \label{qualitative_results_tod}
}
\end{figure*}

\subsection{Pose Estimation Results on TOD}
For evaluation on the TOD dataset~\citep{liu2020keypose}, we compare against KeyPose~\citep{liu2020keypose}, the state-of-the-art method for 6D pose estimation of transparent objects. KeyPose is also the leading approach reported on the TOD dataset. We follow the KeyPose experimental protocol, using the same training and testing data to ensure a fair comparison. In our experiments, we evaluate a total of six objects, selecting one representative object from each category. For each object, one texture is held out for testing, resulting in approximately 3,000 training samples and 320 test samples per object.

For multi-view evaluation, we use four stereo pairs, corresponding to eight viewpoints. KeyPose estimates object poses from each stereo pair independently, and the resulting poses are fused using the CosyPose multi-view fusion procedure to produce globally consistent pose estimates. Table~\ref{tab_keypose} reports the results on the TOD dataset. Our method outperforms ``KeyPose+CosyPose'' on nearly all objects. Although the improvement is modest under the ADD* metric (11.4\%), the gain under the stricter ($5\:\text{mm}, 10^{\circ}$) metric is substantial (32.3\%), highlighting the precision of our approach. Figure~\ref{qualitative_results_tod} provides qualitative examples on different TOD objects, demonstrating the consistency of our pose estimation across diverse scenes.

\begin{table*}[b]
\centering
\resizebox{0.95\textwidth}{!}{
\begin{tabular}{llcccccc}
\toprule
\multicolumn{2}{c}{\multirow{3}{*}{ \:\:\: \:\:\:Dataset \:\:\:\:\:\:\:\:\:\:\:\:\:\: Metric}} 
    & \multicolumn{4}{c}{MV-3D-KP} & Ours \\ 
\cmidrule(lr){3-6} \cmidrule(lr){7-7}
& & RGB+GT Depth* & \multicolumn{2}{c}{RGB+Raw Depth} & RGB+DA+GS & RGB Only \\
\midrule
\multirow{2}{*}{\:\:ROBI (Ensenso)} 
 & ADD*    & 95.5 & \multicolumn{2}{c}{\textbf{94.6}} & 37.5 & \underline{90.9} \\
 & (5,10) & 93.8 & \multicolumn{2}{c}{\textbf{92.0}} & 28.1 & \underline{80.0} \\
\midrule
\multirow{2}{*}{ROBI (RealSense)} 
 & ADD*    & 95.3 & \multicolumn{2}{c}{\underline{89.2}} & 34.9  & \textbf{90.1} \\
 & (5,10) & 93.2 & \multicolumn{2}{c}{\textbf{79.6}} & 31.6  & \underline{76.8} \\
\midrule
\multirow{2}{*}{ \:\:\:\:\:\:\:T-ROBI} 
 & ADD*    & 100  & \multicolumn{2}{c}{33.1} & \underline{60.1}  & \textbf{95.2} \\
 & (5,10) & 100 & \multicolumn{2}{c}{32.2} & \underline{34.1} & \textbf{81.3} \\
\bottomrule
\end{tabular}}
\caption{
Comparison of our RGB-only approach against MV-3D-KP with different depth sources on the ROBI and T-ROBI datasets. \textbf{GT depth*} is an oracle reference (not available in practice). \textbf{Ensenso} represents a high-end depth sensor, \textbf{RealSense} a commodity sensor, and \textbf{DA+GS} refers to Depth-Anything v2 with per-frame GT scale and shift alignment. All results are computed using eight viewpoints per scene. \textbf{Bold} indicates the best practical result, and \underline{underline} indicates the second-best result.}
\label{tab_ablation_depth}
\end{table*}

\subsection{Comparison with RGB-D Baselines}
To evaluate the practical advantage of our RGB-only approach, we compare it against the MV-3D-KP baseline on ROBI and T-ROBI dataset under different depth configurations:
\begin{itemize}
    \item \textbf{Ground-Truth (GT) Depth}: Captured with scanning spray and a high-end Ensenso sensor for optimal quality. This depth serves as an oracle reference and is not available in practice.
    \item \textbf{Raw Depth}: Directly captured from either a high-end Ensenso sensor or a commodity-level RealSense camera. 
    \item \textbf{DA + GT Scale}: Depth is predicted from the RGB image using Depth-Anything V2~\citep{yang2024depth}. For each frame, the predicted relative depth is converted to metric depth by aligning it to the corresponding ground-truth depth using scale and shift~\citep{ganj2025hybriddepth}. This represents the upper-bound performance of Depth-Anything V2.
\end{itemize}
Results are summarized in Table~\ref{tab_ablation_depth}. As expected, MV-3D-KP achieves near-perfect accuracy with GT depth, establishing an upper bound for RGB-D performance. With raw sensor depth on ROBI datasets (reflective objects), MV-3D-KP slightly outperforms our multi-view RGB-only method when using the high-end Ensenso sensor, while with a commodity RealSense sensor, our method achieves comparable performance. On T-ROBI datasets (transparent objects), where a large portion of depth data is missing, our RGB-only approach significantly outperforms MV-3D-KP by at least 49.1\%.

When using Depth-Anything V2, MV-3D-KP performance improves slightly only for objects that are completely invisible to the sensor, such as the transparent objects in T-ROBI. Even in these cases, it still performs worse than our RGB-only method by at least 35.1\%. In contrast, on the ROBI dataset, where objects are reflective but depth can be directly sensed, DA+GT provides minimal improvement and performs worse than on T-ROBI, mainly due to the high clutter in ROBI bins. For these ROBI objects, MV-3D-KP with DA+GT remains far below both its performance with real depth and our RGB-only approach. Overall, these results demonstrate that while predicted depth can help in extreme cases, it cannot replace real depth measurements or multi-view RGB cues for robust pose estimation.

\begin{table*}[t]
\centering
\resizebox{0.95\textwidth}{!}{
\begin{tabular}{cccccccc}
\toprule
\multirow{2}{*}{Object Edge} & \multirow{2}{*}{Sequential Process} 
    & \multicolumn{2}{c}{ADD*} & \multicolumn{2}{c}{($5\,mm$, $10^{\circ}$)} & \multirow{2}{*}{Run-time (ms)} \\ 
\cmidrule(lr){3-4} \cmidrule(lr){5-6}
& & 4 Views & 8 Views & 4 Views & 8 Views & \\
\midrule
$\times$    & $\times$      & 70.3 & 78.4 & 48.3 & 55.4 & 157.2 \\
$\times$    & \checkmark    & 73.3 & 79.7 & 48.5 & 55.8 & \textbf{34.1} \\
\checkmark  & $\times$      & \underline{85.4} & \underline{90.1} & \underline{71.9} & \underline{78.6} & 197.7 \\
\checkmark  & \checkmark    & \textbf{86.5} & \textbf{91.2} & \textbf{72.1} & \textbf{79.3} & \underline{48.5} \\
\bottomrule
\end{tabular}}
\caption{Ablation studies on different configurations for 6D object pose estimation on the ROBI and T-ROBI datasets. Results report detection rate based on ADD* and 5-mm/10-degree metrics. \textbf{Object Edge} refers to utilizing the object's 2D edge map from MEC-Net to obtain per-frame object orientation measurements. \textbf{Sequential Process} shows our method when 6D pose estimation is decomposed into a two-step sequential process. We report run-time for per-frame orientation estimation using LINE-2D in milliseconds per object, tested on a laptop with an Intel 2.60GHz CPU. \textbf{Bold} indicates the best practical result, \underline{underline} the second-best.}
\label{tab_ablation}
\end{table*}

\begin{table*}[b]
\centering
\resizebox{0.6\textwidth}{!}{
\begin{tabular}{lcccccc}
\toprule
\multirow{2}{*}{Dataset} & \multicolumn{3}{c}{ADD*} & \multicolumn{3}{c}{($5\,mm$, $10^{\circ}$)} \\
\cmidrule(lr){2-4} \cmidrule(lr){5-7}
 & Random & Max & NBV & Random & Max & NBV \\
\midrule
ROBI   & 85.5 & \underline{88.3} & \textbf{89.6} & 70.2 & \underline{74.1} & \textbf{76.1} \\
T-ROBI & 95.3 & \underline{96.3} & \textbf{97.2} & 79.4 & \textbf{85.1} & \underline{84.1} \\
TOD    & 97.3  & \textbf{99.7}  & \textbf{99.7}  & 79.5  & \underline{89.3}  & \textbf{90.7} \\
\bottomrule
\end{tabular}}
\caption{Next-Best-View evaluation. We show the object pose estimation results with different viewpoint selection strategies on \textbf{ROBI}, \textbf{T-ROBI} and \textbf{TOD} dataset. An object pose is considered correct if it lies within the ADD* or ($5\:mm$, $10^{\circ}$) metric. We initialize the pose estimation with two viewpoints. The maximum number of additional viewpoints is set to two (a total of four viewpoints).}
\label{tab_nbv}
\end{table*}

\subsection{Ablation Studies on Pose Estimation}
We conduct ablation studies to evaluate the effect of using the edge map and the decoupled formulation on the ROBI and T-ROBI dataset. Table~\ref{tab_ablation} summarizes the results of our ablation studies.

\textbf{Edge Map.} As presented in Section~\ref{Sec:rotation}, for optimizing the object 3D orientation, we use a template matching-based orientation estimator, LINE-2D, to obtain the per-frame object orientation measurement. However, LINE-2D is susceptible to issues related to occlusion and fake edges. Compared to our previous approach~\citep{yang20236d}, we address these problems by leveraging our MEC-Net to directly produce the object's 2D edge map. To demonstrate the advantage of this approach, we conduct a comparison of the final results with and without using the edge map. In cases where edge maps are unavailable, we take the object mask from the MEC-Net and then feed the re-cropped object RoI into the LINE-2D estimator. Table~\ref{tab_ablation} clearly shows a significant increase in the correct detection rate when utilizing the estimated edge map. This phenomenon is more obvious when using the metric, ($5\:mm$, $10^{\circ}$), which imposes a stricter criterion for orientation error.

\textbf{Sequential Process.} As discussed in Section~\ref{sec_pose} and \ref{sec_rot_meas}, the core idea of our method is the decoupling of 6D pose estimation into a two-step sequential process. This process first resolves the scale and depth ambiguities in the RGB images and greatly improves the orientation estimation performance. To justify its effectiveness, we consider an alternative version of our approach, one which simultaneously estimates the 3D translation and orientation. This version uses the same strategy to estimate the object translation. However, instead of using the provided scale from the translation estimates, it uses the multi-scale trained templates (similar to the RGB version of CosyPose) to acquire orientation measurements. Table~\ref{tab_ablation} shows that, due to the large number of templates, the run-time for orientation estimation is generally slow for the simultaneous process version. In comparison, our sequential process not only operates with a much faster run-time speed but also has slightly better overall performance.

\begin{figure*}[t]
\centering
\begin{subfigure}{0.32\textwidth}
  \includegraphics[width=\linewidth]{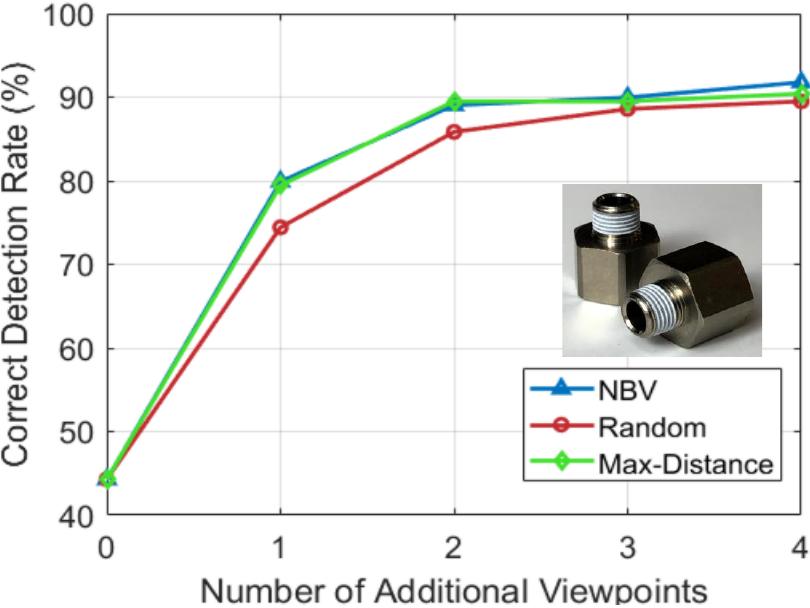}
  \caption{ROBI: Tube Fitting}
\end{subfigure}
\begin{subfigure}{0.32\textwidth}
    \includegraphics[width=\linewidth]{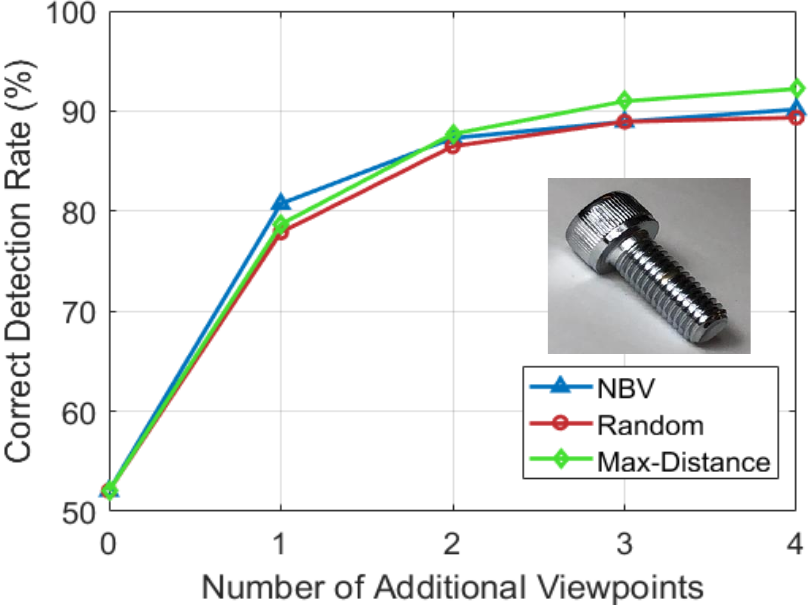}
      \caption{ROBI: Chrome Screw}
\end{subfigure}
\begin{subfigure}{0.32\textwidth}
  \includegraphics[width=\linewidth]{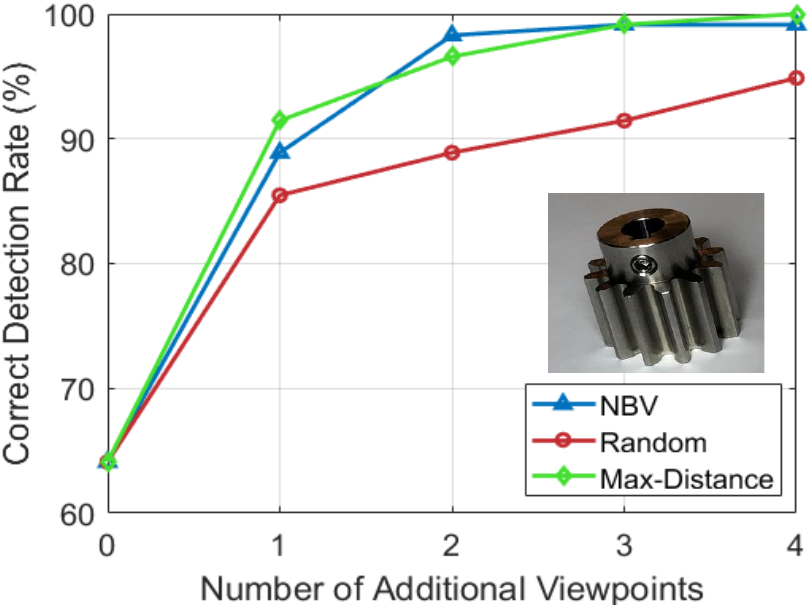}
  \caption{ROBI: Gear}
\end{subfigure}
\begin{subfigure}{0.32\textwidth}
    \includegraphics[width=\linewidth]{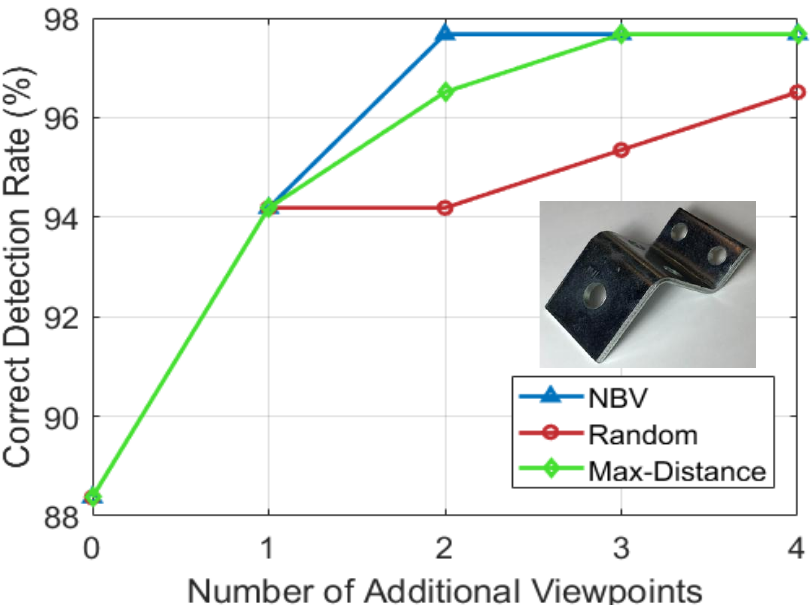}
      \caption{ROBI: Zigzag}
\end{subfigure}
\begin{subfigure}{0.32\textwidth}
    \includegraphics[width=\linewidth]{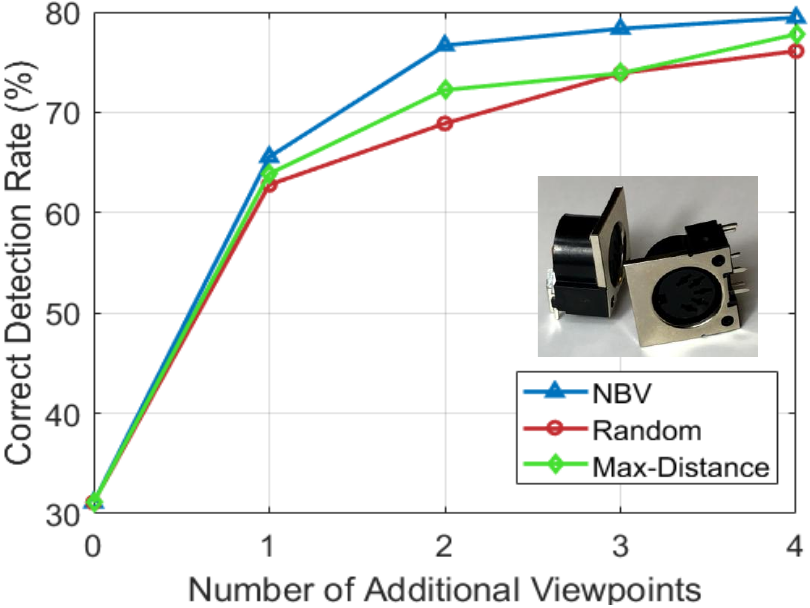}
      \caption{ROBI: DIN Connector}
\end{subfigure}
\begin{subfigure}{0.32\textwidth}
    \includegraphics[width=\linewidth]{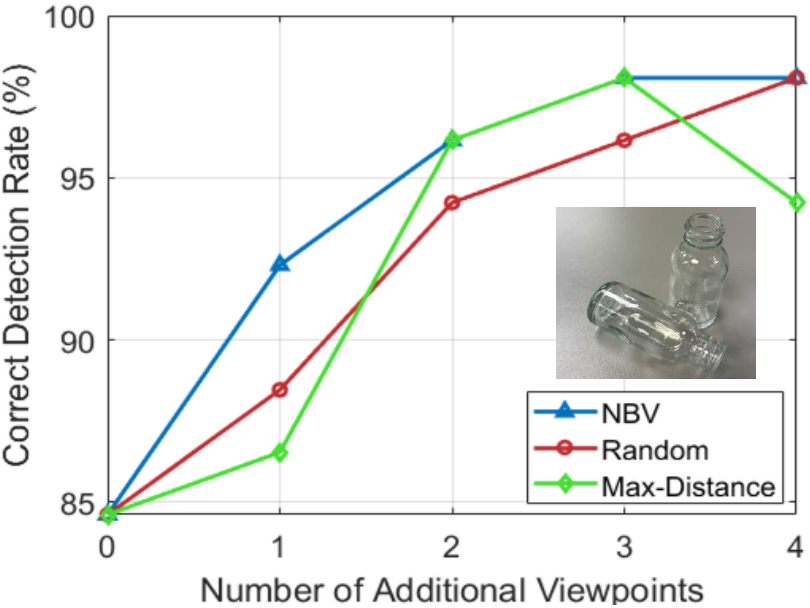}
      \caption{T-ROBI:  Bottle}
\end{subfigure}
\begin{subfigure}{0.32\textwidth}
    \includegraphics[width=\linewidth]{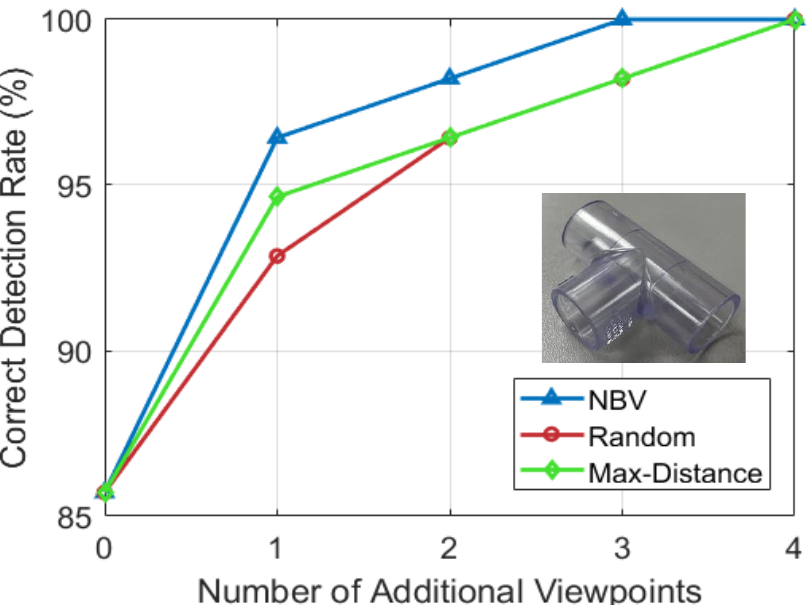}
      \caption{T-ROBI: Pipe Fitting}
\end{subfigure}
\begin{subfigure}{0.32\textwidth}
    \includegraphics[width=\linewidth]{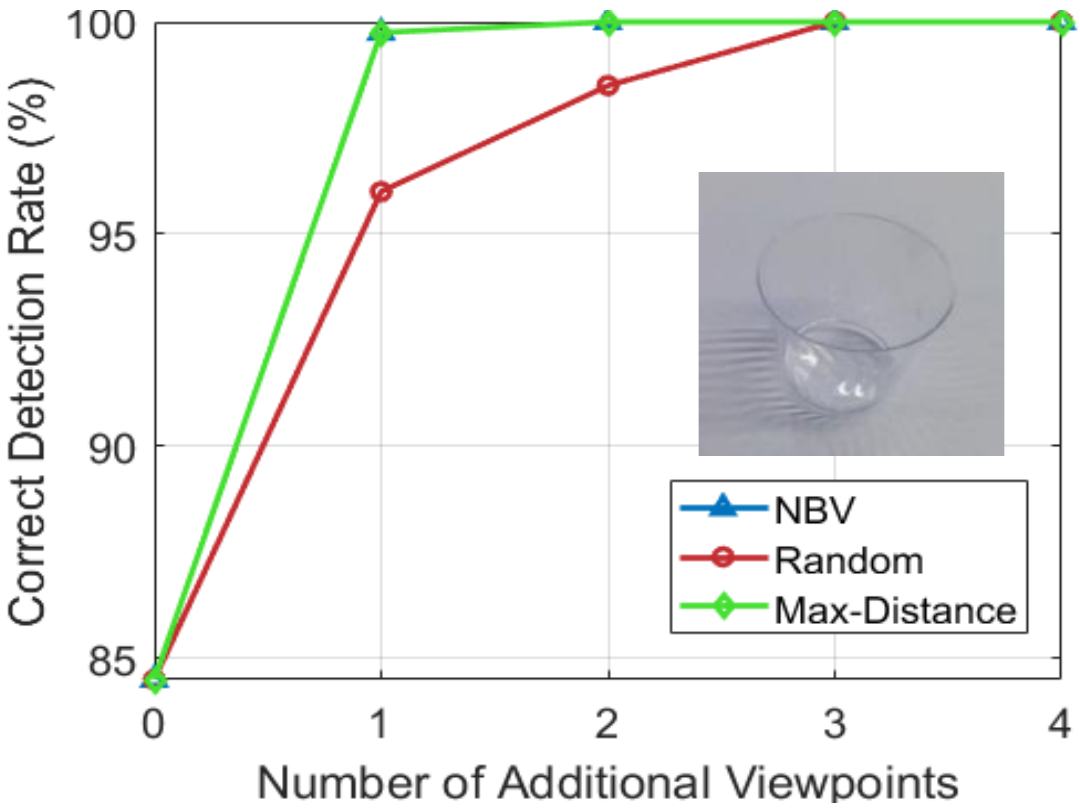}
      \caption{TOD: Cup}
\end{subfigure}
\begin{subfigure}{0.32\textwidth}
    \includegraphics[width=\linewidth]{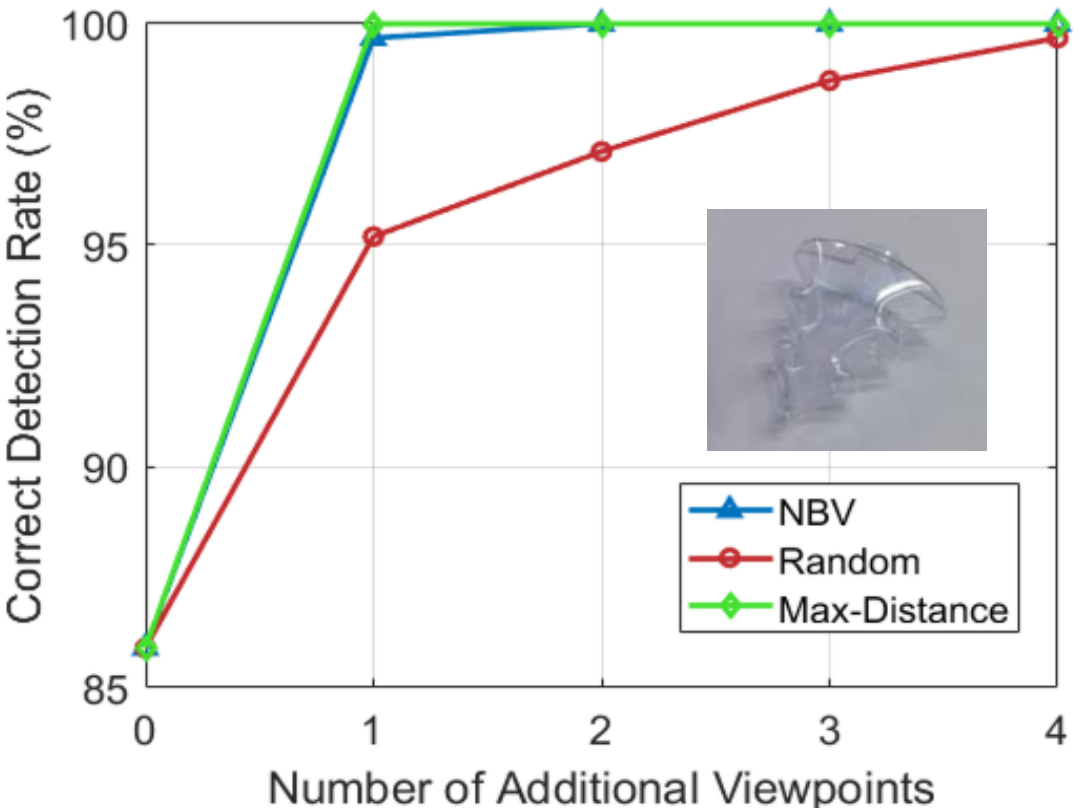}
      \caption{TOD: Tree}
\end{subfigure}
\caption[Evaluation of the NBV policy when comparing against baselines on RGB-based object pose estimation.]{Evaluation of our next-best-view policy when comparing against heuristic-based baselines. We use our multi-view pose estimation approach for all the viewpoint selection strategies. The results are evaluated using the correct detection rate with the \textbf{ADD* metric} on the ROBI, T-ROBI, and TOD datasets. Our approach can achieve a high correct detection rate with fewer viewpoints.}
\label{fig_nbv_results}
\end{figure*}

\subsection{Next-Best-View Evaluation}
In our setup, we operate with a predefined set of camera viewpoints, and the evaluation consists of selecting the next-best viewpoint from this set. We compare our approach against two heuristic-based baselines, ``Random'' and ``Max-Distance''. ``Random'' selects viewpoints randomly from the candidate set, while ``Max-Distance'' moves the camera to the viewpoint farthest from previous observations.

The evaluation is performed on the ROBI, T-ROBI, and TOD datasets. For all view selection strategies, we use our object pose estimation method to ensure a fair comparison. To obtain the results, we initialize the object pose with two viewpoints and progressively refine it by incorporating RGB measurements selected according to each view selection strategy.

Table~\ref{tab_nbv} shows the NBV result when using 2 additional viewpoints (4 viewpoints in total). Our NBV approach outperforms the ``Random'' baseline by a significant margin on all three datasets, with the improvement being more substantial under the stricter ($5,mm$, $10^\circ$) metric, for example, 11.2\% on TOD and 5.9\% on ROBI. Compared to the ``Max-Distance'' baseline, NBV provides smaller yet meaningful improvements, demonstrating that it can select more informative viewpoints and further refine pose estimation beyond heuristic strategies. The benefit of NBV is most evident in highly cluttered scenes such as ROBI, where occlusions are frequent and careful view selection can further enhance pose estimation, improving ADD* by 1.3\% and ($5,mm$, $10^\circ$) by 2\% over the ``Max-Distance'' baseline. Although these improvements are smaller on T-ROBI and TOD, where objects are less cluttered or isolated, NBV still provides slight gains or maintains comparable performance, demonstrating its robustness across different environments.

Figure~\ref{fig_nbv_results} illustrates the trend more clearly when extending to 4 additional viewpoints (6 viewpoints in total). Compared to the ``Random'' (red curve) and ``Max-Distance'' baseline (green curve), our NBV policy (blue curve) consistently achieves higher or comparable ADD* performance on ROBI, T-ROBI, and TOD datasets, showing its reliability across different scene complexities.
\\

\section{Limitations and Future Work}
Although we have demonstrated the effectiveness of our approach in real-world scenes, there are several limitations that future work can address. First, in our problem formulation (Section~\ref{sec_pose}), we model the object translation distribution as a unimodal Gaussian. While this assumption generally holds, it can fail in heavily occluded cases, such as a cylindrical object with both ends occluded, and adopting a multimodal distribution~\citep{bui20206d} could allow the model to capture multiple plausible translations. 

Second, in our next best view prediction, although the object’s orientation is modeled as a multimodal distribution, the predicted viewpoints are intended to refine the final pose accuracy under the assumption that the initial pose is unambiguous (Section~\ref{NBV}). Consequently, if the object exhibits inherent visual ambiguity (e.g., from occlusion), this approach cannot resolve it, since it does not explicitly account for disambiguating between multiple plausible modes~\citep{manhardt2019explaining}. Addressing these ambiguities is an important direction for future work.

Third, in our NBV setup, we assume a predefined set of camera viewpoints, which can be directly mapped to the robot’s poses for execution on a real robot platform but restricts the robot to discrete positions. Generating continuous motions via trajectory optimization~\citep{falanga2018pampc, wang2019manipulation} could enable more informative observations, particularly on platforms equipped with an end-effector-mounted camera.

Finally, our current approach requires a 3D object CAD model and known camera poses, which limits its applicability. Future work will investigate joint estimation of object and camera poses and explore extending the active perception framework to CAD-less objects~\cite{wang2021dsp, liao2024uncertainty}.

\section{Conclusion}
\label{sec6}
In this work, we present a complete framework of multi-view pose estimation and next-best-view prediction for textureless objects. For our multi-view object pose estimation approach, the core idea of our method is to decouple the posterior distribution into a 3D translation and a 3D orientation of an object and integrate the per-frame measurements with a two-step sequential formulation. This process first resolves the scale and depth ambiguities in the RGB images and greatly simplifies the per-frame orientation estimation problem. Moreover, our orientation optimization module explicitly handles the object symmetries and counteracts the measurement uncertainties with a max-mixture-based formulation. To find the next-best-view, we predict the object pose entropy via the Fisher information approximation. The new RGB measurements are collected from the corresponding viewpoint to improve the object pose accuracy. Experiments on public datasets ROBI and TOD, along with our T-ROBI dataset, demonstrate the effectiveness and accuracy compared to the state-of-the-art baselines.

\bibliographystyle{SageH}
\bibliography{bibliography.bib}

\end{document}